\pdfoutput=1
\documentclass[11pt,a4paper]{article}
\usepackage[hyperref]{acl2017}
\usepackage{times}
\usepackage{latexsym}

\usepackage[T2A, T1]{fontenc}
\usepackage[utf8]{inputenc}

\usepackage{url}
\usepackage{graphicx}
\usepackage{multirow}
\usepackage{amsthm}
\usepackage{amssymb}
\usepackage{amsmath}
\usepackage{amsfonts}
\usepackage{color}
\usepackage{mdwlist}
\usepackage{microtype}
\usepackage{tabularx}
\usepackage{booktabs}
\usepackage{subcaption}
\usepackage{color,colortbl}

\definecolor{Gray}{gray}{0.9}

\newcolumntype{b}{X}
\newcolumntype{m}{>{\hsize=.6\hsize}X}
\newcolumntype{s}{>{\hsize=.33\hsize}X}

\makeatletter
\newcommand{\removelatexerror}{\let\@latex@error\@gobble}
\newcommand{\en}{\textsc{en}~}
\newcommand{\attrepel}{\textsc{Attract-Repel} }

\makeatother

\usepackage[russian,english]{babel}

\aclfinalcopy 
\def\aclpaperid{494} 

\title{Morph-fitting: Fine-Tuning Word Vector Spaces \\ with Simple Language-Specific Rules}

\author{ {Ivan Vuli\'c}$^{\mathbf{1}}$, ~ Nikola Mrk\v{s}i\'{c}$^{\mathbf{1}}$, ~ {Roi Reichart}$^{\mathbf{2}}$ ~  \\
\bf{Diarmuid \'{O} S\'{e}aghdha}$^{\mathbf{3}}$, ~ {{Steve Young}}$^{\mathbf{1}}$,  ~ {Anna Korhonen}$^{\mathbf{1}}$  \\
$^{\mathbf{1}}$ University of Cambridge~~~~~~~  
$^{\mathbf{2}}$ Technion, Israel Institute of Technology~~~~~~~
$^{\mathbf{3}}$ Apple Inc. \\
\texttt{\{iv250,nm480,sjy11,alk23\}@cam.ac.uk} \hspace{0.5em} \\
\texttt{doseaghdha@apple.com} \hspace{0.8em} \texttt{roiri@ie.technion.ac.il}
}

\date{}

\begin{document}
\maketitle
\begin{abstract}

Morphologically rich languages accentuate two properties of distributional vector space models: {1)} the difficulty of inducing accurate representations for low-frequency word forms; and {2)} insensitivity to distinct lexical relations that have similar distributional signatures. These effects are detrimental for language understanding systems, which may infer that \emph{inexpensive} is a rephrasing for \emph{expensive} or may not associate \emph{acquire} with \emph{acquires}. In this work, we propose a novel \emph{morph-fitting} procedure which moves past the use of curated semantic lexicons for improving distributional vector spaces. Instead, our method injects morphological constraints generated using simple language-specific rules, pulling \emph{inflectional} forms of the same word close together and pushing \emph{derivational antonyms} far apart. In intrinsic evaluation over four languages, we show that our approach: \textbf{1)} improves low-frequency word estimates; and \textbf{2)} boosts the semantic quality of the entire word vector collection. Finally, we show that {morph-fitted} vectors yield large gains in the downstream task of \emph{dialogue state tracking}, highlighting the importance of morphology for tackling long-tail phenomena in language understanding tasks.



\end{abstract}

\section{Introduction}
\label{s:intro}
Word representation learning has become a research area of central importance in natural language processing (NLP), with its usefulness demonstrated across many application areas such as parsing \cite{Chen:2014emnlp,Johannsen:2015emnlp}, machine translation \cite{Zou:2013emnlp}, and many others \cite{Turian:2010acl,Collobert:2011jmlr}. Most prominent word representation techniques are grounded in the \textit{distributional hypothesis} \cite{Harris:1954}, relying on word co-occurrence information in large textual corpora \cite[i.a.]{Curran:04,Turney:2010jair,Mikolov:2013nips,Mnih:2013nips,Levy:2014acl,Schwartz:2015conll}. 

Morphologically rich languages, in which ``substantial grammatical information\ldots is expressed at word level'' \cite{Tsarfaty:10}, pose specific challenges for NLP. This is not always considered when techniques are evaluated on languages such as English or Chinese, which do not have rich morphology. In the case of distributional vector space models, morphological complexity brings two challenges to the fore: 
\vspace{1mm}

\textbf{1. Estimating Rare Words:} A single lemma can have many different surface realisations. Naively treating each realisation as a separate word leads to sparsity problems and a failure to exploit their shared semantics. On the other hand, lemmatising the entire corpus can obfuscate the differences that exist between different word forms even though they share some aspects of meaning. \vspace{1mm}

\textbf{2. Embedded Semantics:} Morphology can encode semantic relations such as antonymy (e.g. \emph{literate} and \emph{illiterate}, \emph{expensive} and \emph{inexpensive}) or (near-)synonymy (\emph{north}, \emph{northern}, \emph{northerly}). 

\begin{figure}[t]
\centering
\includegraphics[width=1.00\linewidth]{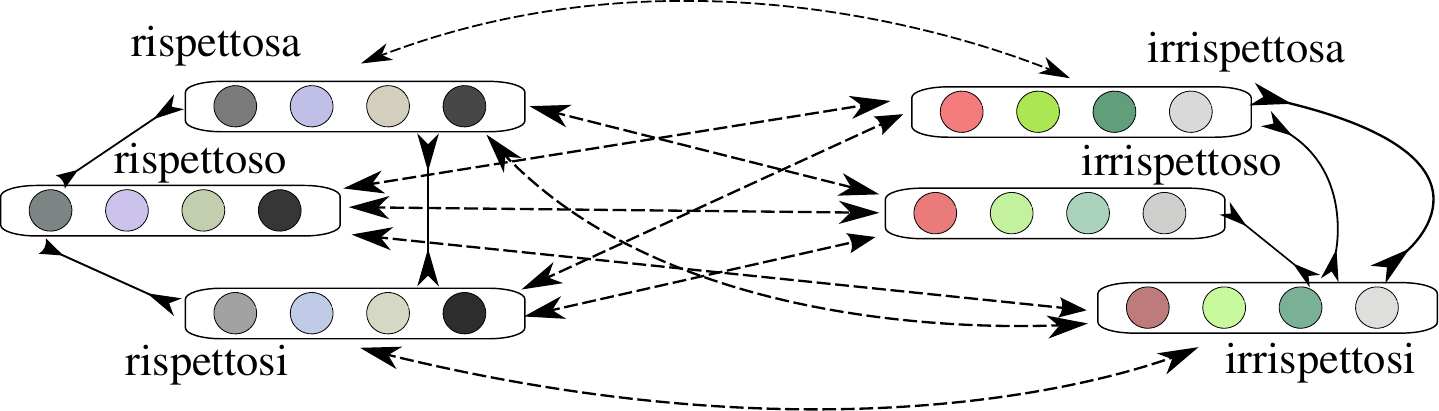}
\vspace{-0.5em}
\caption{\textit{Morph-fitting} in Italian. Representations for \textit{rispettoso}, \textit{rispettosa}, \textit{rispettosi} (\textsc{en}: \textit{respectful}), are pulled closer together in the vector space (solid lines; \textsc{Attract} constraints). At the same time, the model pushes them away from their antonyms (dashed lines; \textsc{Repel} constraints) \textit{irrispettoso}, \textit{irrispettosa}, \textit{irrispettosi} (\textsc{en}: \textit{disrespectful}), obtained through morphological affix transformation captured by language-specific rules (e.g., adding the prefix \textit{ir-} typically negates the base word in Italian)} 
\vspace{-0.8mm}
\label{fig:illustration}
\end{figure}

\begin{table*}
\def\arraystretch{0.98}
\centering
\resizebox{2.05\columnwidth}{!}{%
{\small
\begin{tabular}{ccc|ccc|ccc}
\bf en\_expensive & \bf de\_teure & \bf it\_costoso  & \bf en\_slow & \bf de\_langsam & \bf it\_lento &  \bf en\_book & \bf de\_buch & \bf it\_libro \\ \hline
costly        & teuren & dispendioso & fast & \foreignlanguage{german}{allmählich} & lentissimo &  books & sachbuch & romanzo  \\
costlier      & kostspielige & remunerativo & slowness & rasch & lenta &  memoir & buches & racconto   \\
cheaper       & \foreignlanguage{german}{aufwändige} & redditizio & slower & \foreignlanguage{german}{gemächlich} & inesorabile & novel & \foreignlanguage{german}{romandebüt} & volumetto   \\
prohibitively & kostenintensive & rischioso & slowed & schnell & rapidissimo &  storybooks & \foreignlanguage{german}{büchlein} & saggio   \\
pricey        & aufwendige & costosa & slowing & explosionsartig & graduale &  blurb & pamphlet & ecclesiaste  \\ \hline
expensiveness & teures & costosa &  slowing & langsamer & lenti & booked & \foreignlanguage{german}{bücher} & libri  \\
costly        & teuren & costose &  slowed & langsames & lente & rebook & \foreignlanguage{german}{büch} & libra   \\
costlier      & teurem & costosi &  slowness & langsame & lenta & booking & \foreignlanguage{german}{büche} & librare   \\ 
ruinously     & teurer & dispendioso &  slows & langsamem & veloce & rebooked & \foreignlanguage{german}{büches} & libre  \\
unaffordable  & teurerer & dispendiose &  idle & langsamen & rapido & books & \foreignlanguage{german}{büchen} & librano  \\
\end{tabular}}%
}
\vspace{-0.5em}
\caption{The nearest neighbours of three example words (\emph{expensive}, \emph{slow} and \emph{book}) in English, German and Italian before (top) and after (bottom) { morph-fitting}.  }
\vspace{-0.8mm}
\label{tab:qualitative-table}
\end{table*}
\vspace{1mm}
In this work, we tackle the two challenges jointly by introducing a {\em resource-light} vector space fine-tuning procedure termed \emph{morph-fitting}. The proposed method does not require curated knowledge bases or gold lexicons. Instead, it makes use of the observation that morphology implicitly encodes semantic signals pertaining to synonymy (e.g., German word {inflections} \textit{katalanisch, katalanischem, katalanischer} denote the same semantic concept in different grammatical roles), and antonymy (e.g., {\em mature} vs. {\em immature}), capitalising on the proliferation of word forms in morphologically rich languages. Formalised as an instance of the post-processing \textit{semantic specialisation} paradigm \cite{Faruqui:2015naacl,Mrksic:2016naacl}, {morph-fitting} is steered by a set of linguistic constraints derived from simple language-specific rules which describe (a subset of) morphological processes in a language. The constraints emphasise similarity on one side (e.g., by extracting \textit{morphological} synonyms), and antonymy on the other (by extracting \textit{morphological} antonyms), see Fig.~\ref{fig:illustration} and Tab.~\ref{tab:cons}.

The key idea of the fine-tuning process is to pull synonymous examples described by the constraints closer together in the transformed vector space, while at the same time pushing antonymous examples away from each other. The explicit post-hoc injection of morphological constraints enables: {\bf a)} the estimation of more accurate vectors for low-frequency words which are linked to their high-frequency forms by the constructed constraints;\footnote{For instance, the vector for the word \textit{katalanischem}  which occurs only 9 times in the German Wikipedia will be pulled closer to the more reliable vectors for \textit{katalanisch} and \textit{katalanischer}, with frequencies of 2097 and 1383 respectively.} this tackles the data sparsity problem; and {\bf b)} specialising the distributional space to distinguish between similarity and relatedness \cite{Kiela:2015emnlp}, thus supporting language understanding applications such as {\em dialogue state tracking} (DST).\footnote{Representation models that do not distinguish between synonyms and antonyms may have grave implications in downstream language understanding applications such as spoken dialogue systems: a user looking for {\em `an affordable Chinese restaurant in west Cambridge'} does not want a recommendation for {\em `an expensive Thai place in east Oxford'}. }

As a post-processor, {morph-fitting} allows the integration of morphological rules with any distributional vector space in any language: it treats an input distributional word vector space as a black box and fine-tunes it so that the transformed space reflects the knowledge coded in the input morphological constraints (e.g., Italian words \textit{rispettoso} and \textit{irrispetosa} should be far apart in the transformed vector space, see Fig.~\ref{fig:illustration}). Tab.~\ref{tab:qualitative-table} illustrates the effects of {morph-fitting} by qualitative examples in three languages: the vast majority of nearest neighbours are ``morphological'' synonyms.

We demonstrate the efficacy of {morph-fitting} in four languages (English, German, Italian, Russian), yielding large and consistent improvements on benchmarking word similarity evaluation sets such as SimLex-999 \cite{Hill:2015cl}, its multilingual extension \cite{Leviant:2015arxiv}, and SimVerb-3500 \cite{Gerz:2016emnlp}. The improvements are reported for all four languages, and with a variety of input distributional spaces, verifying the robustness of the approach.

We then show that incorporating {morph-fitted} vectors into a state-of-the-art neural-network DST model results in improved tracking performance, especially for morphologically rich languages. We report an improvement of 4\% on Italian, and 6\% on German when using \textit{morph-fitted} vectors instead of the distributional ones, setting a new state-of-the-art DST performance for the two datasets.\footnote{There are no readily available DST datasets for Russian.}

\section{Morph-fitting: Methodology}
\label{s:metho}
\noindent \textbf{Preliminaries }
In this work, we focus on four languages with varying levels of morphological complexity:  English (\textsc{en}), German (\textsc{de}), Italian (\textsc{it}), and Russian (\textsc{ru}). These correspond to languages in the Multilingual SimLex-999 dataset. Vocabularies $W_{en}$, $W_{de}$, $W_{it}$, $W_{ru}$ are compiled by retaining all word forms from the four Wikipedias with word frequency over 10, see Tab.~\ref{tab:stats}. We then extract sets of linguistic constraints from these (large) vocabularies using a set of simple language-specific \textit{if-then-else} rules, see Tab.~\ref{tab:cons}.\footnote{A native speaker can easily come up with these sets of morphological rules (or at least with a reasonable subset of them) without any linguistic training. What is more, the rules for \textsc{de}, \textsc{it}, and \textsc{ru} were created by non-native, non-fluent speakers with a limited knowledge of the three languages, exemplifying the simplicity and portability of the approach.} These constraints (Sect.~\ref{ss:rules}) are used as input for the vector space post-processing \textsc{Attract-Repel} algorithm (outlined in Sect.~\ref{ss:attrpl}).

\subsection{The \textsc{Attract-Repel} Model}
\label{ss:attrpl}

The \textsc{Attract-Repel} model, proposed by \newcite{Mrksic:17}, is an extension of the \textsc{Paragram} procedure proposed by \newcite{Wieting:2015tacl}. It provides a generic framework for incorporating \emph{similarity} (e.g.~\emph{successful} and \emph{accomplished}) and \emph{antonymy} constraints (e.g.~\emph{nimble} and \emph{clumsy}) into pre-trained word vectors. Given the initial vector space and collections of \textsc{Attract} and \textsc{Repel} constraints $A$ and $R$, the model gradually modifies the space to bring the designated word vectors closer together or further apart. The method's cost function consists of three terms. The first term pulls the \textsc{Attract} examples $(x_l, x_r) \in A$ closer together. If $B_{A}$ denotes the current mini-batch of \textsc{Attract} examples, this term can be expressed as: 

\vspace{-0.0em}
{\small
\begin{eqnarray*} A(\mathcal{B}_A) ~=  \sum_{ (x_l, x_r) \in \mathcal{B}_{A}} &( ReLU\left( \delta_{att} +  \mathbf{x}_l \mathbf{t}_l - \mathbf{x}_l \mathbf{x}_r \right)  \\
+&ReLU\left( \delta_{att} +  \mathbf{x}_r \mathbf{t}_r - \mathbf{x}_l \mathbf{x}_r  \right) )  
\end{eqnarray*}}%

\noindent where $\delta_{att}$ is the similarity margin which determines how much closer synonymous vectors should be to each other than to each of their respective negative examples. $ReLU(x)=\max(0,x)$ is the standard rectified linear unit \cite{Nair:2010icml}. The `negative' example $\mathbf{t}_i$ for each word $x_i$ in any \textsc{Attract} pair is the word vector \emph{closest} to $\mathbf{x}_i$ among the examples in the current mini-batch (distinct from its target synonym and $\mathbf{x}_i$ itself). This means that this term forces synonymous words from the in-batch \textsc{Attract} constraints to be closer to one another than to any other  word in the current mini-batch.  

The second term pushes antonyms away from each other. If $(x_l, x_r) \in  B_{R}$ is the current mini-batch of \textsc{Repel} constraints, this term can be expressed as follows:

{\small
\begin{eqnarray*} R(\mathcal{B}_R) ~= \sum_{(x_l, x_r) \in \mathcal{B}_{R}}  &( ReLU\left(\delta_{rpl} + \mathbf{x}_l \mathbf{x}_r  - \mathbf{x}_l \mathbf{t}_r \right) \\
+&ReLU\left( \delta_{rpl} + \mathbf{x}_l \mathbf{x}_r  - \mathbf{x}_r \mathbf{t}_r  \right)  ) 
\end{eqnarray*}}%
In this case, each word's `negative' example is the (in-batch) word vector furthest away from it (and distinct from the word's target antonym). The intuition is that we want antonymous words from the input \textsc{Repel} constraints to be \emph{further away} from each other than from any other word in the current mini-batch; $\delta_{rpl}$ is now the {\em repel} margin. 

The final term of the cost function serves to retain the abundance of semantic information encoded in the starting distributional space. If $\mathbf{x}_{i}^{init}$ is the initial distributional vector and $V(\mathcal{B})$ is the set of all vectors present in the given mini-batch, this term (per mini-batch) is expressed as follows:

\vspace{-0.5em}
{\small
\begin{equation*}
  R(\mathcal{B}_A, \mathcal{B}_R) =  \sum\limits_{ \mathbf{x}_i \in V(\mathcal{B}_A \cup \mathcal{B}_R) }  \lambda_{reg} \left\| \mathbf{x}^{init}_{i} - \mathbf{x}_i \right\|_{2} 
 \end{equation*}}%
\noindent where $\lambda_{reg}$ is the L2 regularisation constant.\footnote{We use hyperparameter values $\delta_{att}=0.6$, $\delta_{rpl}=0.0$, $\lambda_{reg}=10^{-9}$ from prior work without fine-tuning. We train all models for 10 epochs with AdaGrad \cite{Duchi:11}.} This term effectively \emph{pulls} word vectors towards their initial (distributional) values, ensuring that relations encoded in initial vectors persist as long as they do not contradict the newly injected ones.

\begin{table}[t]
\centering
\vspace{-0.0em}
\def\arraystretch{0.98}
{\scriptsize
\begin{tabularx}{0.48\textwidth}{lll}
\toprule
{\bf English} & {\bf German} & {\bf Italian}\\
\midrule
(discuss, discussed) & (schottisch, schottischem) & (golfo, golfi) \\
(laugh, laughing) & (damalige, damaligen) & (minato, minata) \\
(pacifist, pacifists) & (kombiniere, kombinierte) & (mettere, metto) \\
(evacuate, evacuated) & (schweigt, schweigst) & (crescono, cresci) \\
(evaluate, evaluates) & (hacken, gehackt) & (crediti, credite) \\
\midrule
(dressed, undressed) & (stabil, unstabil) & (abitata, inabitato) \\
(similar, dissimilar) & (geformtes, ungeformt) & (realtà, irrealtà) \\
(formality, informality) & (relevant, irrelevant) & (attuato, inattuato) \\
\bottomrule
\end{tabularx}
}
\vspace{-0.5em}
\caption{Example synonymous (inflectional; top) and antonymous (derivational; bottom) constraints.}
\label{tab:cons}
\vspace{-0.5mm}
\end{table}

\begin{table}
\centering
\vspace{-0.0em}
\def\arraystretch{0.92}
{\footnotesize
\begin{tabularx}{\linewidth}{X XXX}
\toprule
{} & {|$W$|} & {|$A$|} & {|$R$|}\\
\cmidrule(lr){2-4}
{English} & {1,368,891} & {231,448} & {45,964} \\
{German} & {1,216,161} & {648,344} & {54,644} \\
{Italian} & {541,779} & {278,974} & {21,400} \\
{Russian} & {950,783} & {408,400} & {32,174} \\

\bottomrule
\end{tabularx}
}
\vspace{-0.5em}
\caption{Vocabulary sizes and counts of \textsc{Attract} ($A$) and \textsc{Repel} ($R$) constraints.}
\vspace{-2mm}
\label{tab:stats}
\end{table}

\subsection{Language-Specific Rules and Constraints}
\label{ss:rules}
\paragraph{Semantic Specialisation with Constraints} 
The fine-tuning \attrepel procedure is entirely driven by the input \textsc{Attract} and \textsc{Repel} sets of constraints. These can be extracted from a variety of semantic databases such as WordNet \cite{Fellbaum:1998wn}, the Paraphrase Database \cite{Ganitkevitch:2013naacl,Pavlick:2015acl}, or BabelNet \cite{Navigli:12,Ehrmann:14} as done in prior work \cite[i.a.]{Faruqui:2015naacl,Wieting:2015tacl,Mrksic:2016naacl}. In this work, we investigate another option: extracting constraints \textit{without} curated knowledge bases in a spectrum of languages by exploiting inherent language-specific properties related to linguistic morphology. This relaxation ensures a wider portability of \attrepel to languages and domains without readily available or adequate resources.



\paragraph{Extracting \textsc{Attract} Pairs}
The core difference between {\em inflectional} and \textit{derivational morphology} can be summarised in a few lines as follows: the former refers to a set of processes through which the word form expresses meaningful syntactic information, e.g., verb tense, without any change to the semantics of the word. On the other hand, the latter refers to the formation of new words with semantic shifts in meaning \cite{Schone:2001naacl,Haspelmath:2013book,Lazaridou:2013acl,Zeller:2013acl,Cotterell:2017tacl}.

For the \textsc{Attract} constraints, we focus on {\em inflectional} rather than on {\em derivational morphology} rules as the former preserve the full meaning of a word, modifying it only to reflect grammatical roles such as verb tense or case markers (e.g., {\em (en\_read, en\_reads}) or {\em (de\_katalanisch, de\_katalanischer)}). This choice is guided by our intent to fine-tune the original vector space in order to improve the embedded semantic relations. 


We define two rules for English, widely recognised as morphologically simple \cite{Avramidis:2008acl,Cotterell:2016acl}. These are: \textbf{(R1)} \emph{if} $w_1, w_2 \in W_{en}$, where $w_2 = w_1$ + \textit{ing/ed/s}, \emph{then} add $(w_1, w_2)$ and $(w_2, w_1)$ to the set of \textsc{Attract} constraints $A$. This rule yields pairs such as \textit{(look, looks), (look, looking), (look, looked)}. 

If $w[:-1]$ is a function which strips the last character from word $w$, the second rule is: \textbf{(R2)} \emph{if} $w_1$ ends with the letter \textit{e} \textit{and} $w_1 \in W_{en}$ \textit{and} $w_2 \in W_{en}$, where $w_2 = w_1[:-1]$ + \textit{ing/ed}, \emph{then} add $(w_1, w_2)$ and $(w_2, w_1)$ to $A$. This creates pairs such as \textit{(create, creating)} and \textit{(create, created)}. Naturally, introducing more sophisticated rules is possible in order to cover for other special cases and morphological irregularities (e.g., {\em sweep / swept}), but in all our \en experiments, $A$ is based on the two simple \en rules R1 and R2. 

The other three languages, with more complicated morphology, yield a larger number of rules. In Italian, we rely on the sets of rules spanning: (1) regular formation of plural ({\em libro / libri}); (2) regular verb conjugation ({\em aspettare / aspettiamo}); (3) regular formation of past participle ({\em aspettare / aspettato}); and (4) rules regarding grammatical gender ({\em bianco / bianca}). Besides these, another set of rules is used for German and Russian: (5) regular declension (e.g., {\em asiatisch / asiatischem}).   



\paragraph{Extracting \textsc{Repel} Pairs}
As another source of implicit semantic signals, $W$ also contains words which represent \textit{derivational antonyms}: e.g., two words that denote concepts with opposite meanings, generated through a derivational process. We use a standard set of \en ``antonymy'' prefixes: $AP_{en}=$ \textit{\{dis, il, un, in, im, ir, mis, non, anti\}} \cite{Fromkin:2013book}. \emph{If} $w_1, w_2 \in W_{en}$, where $w_2$ is generated by adding a prefix from $AP_{en}$ to $w_1$, \emph{then} $(w_1, w_2)$ and $(w_2, w_1)$ are added to the set of \textsc{Repel} constraints $R$. This rule generates pairs such as \textit{(advantage, disadvantage)} and \emph{(regular, irregular)}. An additional rule replaces the suffix \textit{-ful} with \textit{-less}, extracting antonyms such as \textit{(careful, careless)}.

Following the same principle, we use $AP_{de}=$ \textit{\{un, nicht, anti, ir, in, miss\}}, $AP_{it}=$ \textit{\{in, ir, im, anti\}}, and $AP_{ru}=$ \textit{\{}\foreignlanguage{russian}{не}, \foreignlanguage{russian}{анти}\textit{\}}. For instance, this generates an \textsc{it} pair \textit{(rispettoso, irrispettoso)} (see Fig.~\ref{fig:illustration}). For \textsc{de}, we use another rule targeting suffix replacement: \textit{-voll} is replaced by \textit{-los}.


We further expand the set of \textsc{Repel} constraints by transitively combining antonymy pairs from the previous step with inflectional \textsc{Attract} pairs. This step yields additional constraints such as \textit{(rispettosa, irrispettosi)} (see Fig.~\ref{fig:illustration}). The final $A$ and $R$ constraint counts are given in Tab.~\ref{tab:stats}. The full sets of rules are available as supplemental material.

\section{Experimental Setup}
\label{s:exp}
\paragraph{Training Data and Setup} 
For each of the four languages we train the skip-gram with negative sampling (SGNS) model \cite{Mikolov:2013nips} on the latest Wikipedia dump of each language. We induce 300-dimensional word vectors, with the frequency cut-off set to 10. The vocabulary sizes $|W|$ for each language are provided in Tab.~\ref{tab:stats}.\footnote{Other SGNS parameters were set to standard values \cite{Baroni:2014acl,Vulic:2016acl}: $15$ epochs, $15$ negative samples, global learning rate: $.025$, subsampling rate: $1e-4$. Similar trends in results persist with $d=100,500$.} We label these collections of vectors \textsc{sgns-large}.

\paragraph{Other Starting Distributional Vectors}
We also analyse the impact of \textit{morph-fitting} on other collections of well-known \textsc{en} word vectors. These vectors have varying vocabulary coverage and are trained with different architectures. We test standard distributional models: Common-Crawl GloVe \cite{Pennington:2014emnlp}, SGNS vectors \cite{Mikolov:2013nips} with various contexts (\textit{BOW} = bag-of-words; \textit{DEPS} = dependency contexts), and training data (\textit{PW} = Polyglot Wikipedia from \newcite{AlRfou:2013conll}; \textit{8B} = 8 billion token \texttt{word2vec} corpus), following \cite{Levy:2014acl} and \cite{Schwartz:2015conll}. We also test the symmetric-pattern based vectors of \newcite{Schwartz:2016naacl} ({\it SymPat-Emb}), count-based PMI-weighted vectors reduced by SVD \cite{Baroni:2014acl} ({\it Count-SVD}), a model which replaces the  context modelling function from CBOW with bidirectional LSTMs \cite{Melamud:2016conll} ({\it Context2Vec}), and two sets of \textsc{en} vectors trained by injecting multilingual information: \textit{BiSkip} \cite{Luong:2015naacl} and \textit{MultiCCA} \cite{Faruqui:2014eacl}. 

We also experiment with standard well-known distributional spaces in other languages (\textsc{it} and \textsc{de}), available from prior work \cite{Dinu:2015arxiv,Luong:2015naacl,Vulic:2016acluniversal}. 

\paragraph{{Morph-fixed} Vectors} 
A baseline which utilises an equal amount of knowledge as {morph-fitting}, termed \textit{morph-fixing}, {fixes} the vector of each word to the distributional vector of its most frequent inflectional synonym, tying the vectors of low-frequency words to their more frequent inflections. For each word $w_1$, we construct a set of $M+1$ words $W_{w_1}=\{w_1,w'_1,\ldots,w'_M\}$ consisting of the word $w_1$ itself and all $M$ words which co-occur with $w_1$ in the \textsc{Attract} constraints. We then choose the word $w'_{max}$ from the set $W_{w_1}$ with the maximum frequency in the training data, and fix all other word vectors in $W_{w_1}$ to its word vector. The {morph-fixed} vectors (\textsc{MFix}) serve as our primary baseline, as they outperformed another straightforward baseline based on \textit{stemming} across all of our intrinsic and extrinsic experiments.

\paragraph{Morph-fitting Variants} 
We analyse two variants of morph-fitting: (1) using \textsc{Attract} constraints only (\textsc{MFit-A}), and (2) using both \textsc{Attract} and \textsc{Repel} constraints (\textsc{MFit-AR}).

\section{Intrinsic Evaluation: Word Similarity}
\label{s:intrinsic}
\paragraph{Evaluation Setup and Datasets}
The first set of experiments intrinsically evaluates \textit{morph-fitted} vector spaces on word similarity benchmarks, using Spearman's rank correlation as the evaluation metric. First, we use the SimLex-999 dataset, as well as SimVerb-3500, a recent \textsc{en} verb pair similarity dataset providing similarity ratings for 3,500 verb pairs.\footnote{Unlike other gold standard resources such as WordSim-353 \cite{Finkelstein:2002tois} or MEN \cite{Bruni:2014jair}, SimLex and SimVerb provided explicit guidelines to discern between semantic similarity and association, so that related but non-similar words (e.g. {\em cup} and {\em coffee}) have a low rating.} SimLex-999 was translated to \textsc{de}, \textsc{it}, and \textsc{ru} by \newcite{Leviant:2015arxiv}, and they crowd-sourced similarity scores from native speakers. We use this dataset for our multilingual evaluation.\footnote{Since \newcite{Leviant:2015arxiv} re-scored the original \textsc{en} SimLex, we use their \textsc{en} SimLex version for consistency.}

\paragraph{Morph-fitting \textsc{en} Word Vectors}
As the first experiment, we {morph-fit} a wide spectrum of \textsc{en} distributional vectors induced by various architectures (see Sect.~\ref{s:exp}). The results on SimLex and SimVerb are summarised in Tab.~\ref{tab:11vectors}. The results with \textsc{en} \textsc{sgns-large} vectors are shown in Fig.~\ref{fig:en}. Morph-fitted vectors bring consistent improvement across all experiments, regardless of the quality of the initial distributional space. This finding confirms that the method is robust: its effectiveness does not depend on the architecture used to construct the initial space. To illustrate the improvements, note that the best score on SimVerb for a model trained on running text is achieved by \textit{Context2vec} ($\rho=0.388$); injecting morphological constraints into this vector space results in a gain of $7.1$ $\rho$ points. 
\begin{table}[t]
\centering
\def\arraystretch{0.92}
\vspace{-0.0em}
{\footnotesize
\begin{tabularx}{\linewidth}{l l l}
\toprule
{} & \multicolumn{2}{l}{\bf Evaluation} \\
\cmidrule(lr){2-3} 
{\bf Vectors} & {SimLex-999} & {SimVerb-3500}  \\
\cmidrule(lr){2-2} \cmidrule(lr){3-3}
{1. SG-BOW2-PW (300)} & {} & {} \\ 
{\scriptsize \cite{Mikolov:2013nips}} & {.339 $\rightarrow$ \textbf{.439}} & {.277 $\rightarrow$ \textbf{.381}} \\
{2. GloVe-6B (300)} & {} & {} \\ 
{\scriptsize \cite{Pennington:2014emnlp}} & {.324 $\rightarrow$ \textbf{.438}} & {.286 $\rightarrow$ \textbf{.405}} \\
{3. Count-SVD (500)} & {} & {} \\ 
{\scriptsize \cite{Baroni:2014acl}} & {.267 $\rightarrow$ \textbf{.360}} & {.199 $\rightarrow$ \textbf{.301}} \\
{4. SG-DEPS-PW (300)} & {} & {} \\ 
{\scriptsize \cite{Levy:2014acl}} & {.376 $\rightarrow$ \textbf{.434}} & {.313 $\rightarrow$ \textbf{.418}} \\
{5. SG-DEPS-8B (500)} & {} & {} \\ 
{\scriptsize \cite{Bansal:2014acl}} & {.373 $\rightarrow$ \textbf{.441}} & {.356 $\rightarrow$ \textbf{.473}} \\
{6. MultiCCA-EN (512)} & {} & {} \\ 
{\scriptsize \cite{Faruqui:2014eacl}} & {.314 $\rightarrow$ \textbf{.391}} & {.296 $\rightarrow$ \textbf{.354}} \\
{7. BiSkip-EN (256)} & {} & {} \\ 
{\scriptsize \cite{Luong:2015naacl}} & {.276 $\rightarrow$ \textbf{.356}} & {.260 $\rightarrow$ \textbf{.333}} \\
{8. SG-BOW2-8B (500)} & {} & {} \\ 
{\scriptsize \cite{Schwartz:2015conll}} & {.373 $\rightarrow$ \textbf{.440}} & {.348 $\rightarrow$ \textbf{.441}} \\
{9. SymPat-Emb (500)} & {} & {} \\ 
{\scriptsize \cite{Schwartz:2016naacl}} & {.381 $\rightarrow$ \textbf{.442}} & {.284 $\rightarrow$ \textbf{.373}} \\
{10. Context2Vec (600)} & {} & {} \\ 
{\scriptsize \cite{Melamud:2016conll}} & {.371 $\rightarrow$ \textbf{.440}} & {.388 $\rightarrow$ \textbf{.459}} \\
\bottomrule
\end{tabularx}
}
\vspace{-0.3em}
\caption{The impact of {morph-fitting} (\textsc{MFit-AR} used) on a representative set of \textsc{en} vector space models. All results show the Spearman's $\rho$ correlation before and after {morph-fitting}. The numbers in parentheses refer to the vector dimensionality.}
\vspace{-0.5mm}
\label{tab:11vectors}
\end{table}

\begin{table}[t]
\centering
\def\arraystretch{0.9}
\vspace{-0.0em}
{\footnotesize
\begin{tabularx}{\linewidth}{l Xll}
\toprule
{\bf Vectors} & {\scriptsize Distrib.} & {\scriptsize \textsc{MFit-A}} & {\scriptsize \textsc{MFit-AR}}  \\
\cmidrule(lr){2-2} \cmidrule(lr){3-3} \cmidrule(lr){4-4}
{\textsc{en}: GloVe-6B (300)} & {.324} & {.376} & {\bf .438} \\
{\textsc{en}: SG-BOW2-PW (300)} & {.339} & {.385} & {\bf .439} \\
\midrule
{\textsc{de}: SG-DEPS-PW (300)} & {} & {} & {} \\
{\scriptsize \cite{Vulic:2016acluniversal}} & {.267} & {.318} & {\bf .325} \\
{\textsc{de}: BiSkip-DE (256)} & {} & {} & {} \\
{\scriptsize \cite{Luong:2015naacl}} & {.354} & {.414} & {\bf .421} \\
\midrule
{\textsc{it}: SG-DEPS-PW (300)} & {} & {} & {} \\
{\scriptsize \cite{Vulic:2016acluniversal}} & {.237} & {.351} & {\bf .391} \\
{\textsc{it}: CBOW5-Wacky (300)} & {} & {} & {} \\
{\scriptsize \cite{Dinu:2015arxiv}} & {.363} & {.417} & {\bf .446} \\

\bottomrule
\end{tabularx}
}
\vspace{-0.3em}
\caption{Results on multilingual SimLex-999 (\textsc{en}, \textsc{de}, and \textsc{it}) with two morph-fitting variants.}
\vspace{-1.1mm}
\label{tab:foreign-vectors}
\end{table}




\paragraph{Experiments on Other Languages} 
We next extend our experiments to other languages, testing both {morph-fitting} variants. The results are summarised in Tab.~\ref{tab:foreign-vectors}, while Fig.~\ref{fig:en}-\ref{fig:ru} show results for the {morph-fitted} \textsc{sgns-large} vectors. These scores confirm the effectiveness and robustness of {morph-fitting} across languages, suggesting that the idea of fitting to morphological constraints is indeed language-agnostic, given the set of language-specific rule-based constraints. Fig.~\ref{fig:main} also demonstrates that the {morph-fitted} vector spaces consistently outperform the {morph-fixed} ones. 

\begin{figure}[t]
\centering
\includegraphics[width=1.01\linewidth]{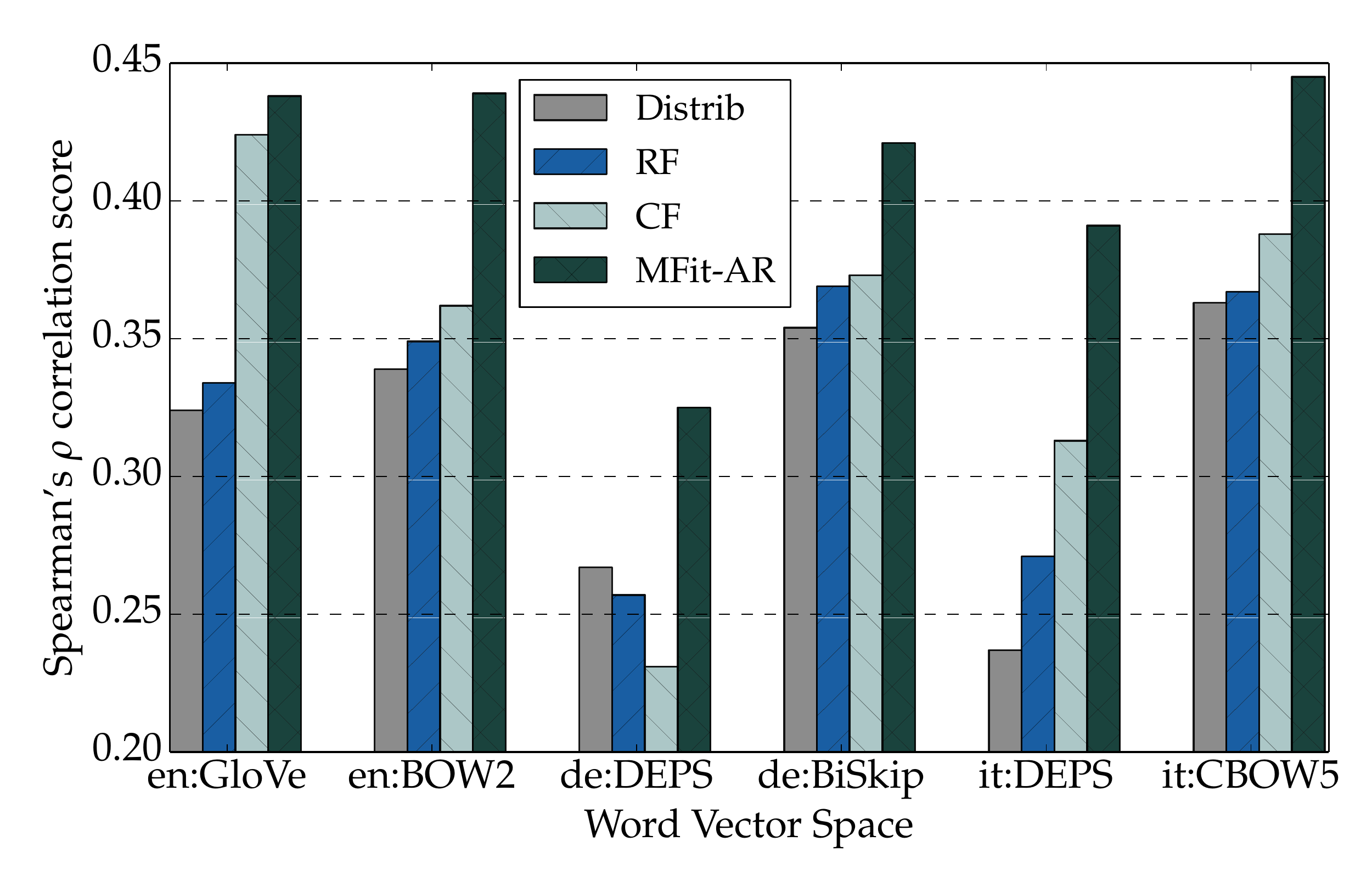}
\vspace{-0.8em}
\caption{A comparison of morph-fitting (the \textsc{MFit-AR} variant) with two other standard specialisation approaches using the same set of morphological constraints: Retrofitting (RF) \cite{Faruqui:2015naacl} and Counter-fitting (CF) \cite{Mrksic:2016naacl}. Spearman's $\rho$ correlation scores on the multilingual SimLex-999 dataset for the same six distributional spaces from Tab.~\ref{tab:foreign-vectors}.}
\vspace{-1.6mm}
\label{fig:arrfcf}
\end{figure}


The comparison between \textsc{MFit-A} and \textsc{MFit-AR} indicates that both sets of constraints are important for the fine-tuning process. \textsc{MFit-A} yields consistent gains over the initial spaces, and (consistent) further improvements are achieved by also incorporating the antonymous \textsc{Repel} constraints. This demonstrates that both types of constraints are useful for semantic specialisation.

\paragraph{Comparison to Other Specialisation Methods}
We also tried using other post-processing specialisation models from the literature in lieu of \textsc{Attract-Repel} using the same set of ``morphological'' synonymy and antonymy constraints. We compare \textsc{Attract-Repel} to the retrofitting model of \cite{Faruqui:2015naacl} and counter-fitting \cite{Mrksic:16b}. The two baselines were trained for 20 iterations using suggested settings. The results for \textsc{en}, \textsc{de}, and \textsc{it} are summarised in Fig.~\ref{fig:arrfcf}. They clearly indicate that \textsc{MFit-AR} outperforms the two other post-processors for each language. We hypothesise that the difference in performance mainly stems from context-sensitive vector space updates performed by \textsc{Attract-Repel}. Conversely, the other two models perform pairwise updates which do not consider what effect each update has on the example pair's relation to other word vectors (for a detailed comparison, see \cite{Mrksic:17}). 

Besides their lower performance, the two other specialisation models have additional disadvantages compared to the proposed morph-fitting model. First, retrofitting is able to incorporate only synonymy/\textsc{Attract} pairs, while our results demonstrate the usefulness of both types of constraints, both for intrinsic evaluation (Tab.~\ref{tab:foreign-vectors}) and downstream tasks (see later Fig.~\ref{fig:main}). Second, counter-fitting is computationally intractable with \textsc{sgns-large} vectors, as its regularisation term involves the computation of all pairwise distances between words in the vocabulary.





\paragraph{Further Discussion} The simplicity of the used language-specific rules does come at a cost of occasionally generating incorrect linguistic constraints such as \textit{(tent, intent)}, \textit{(prove, improve)} or \textit{(press, impress)}. In future work, we will study how to further refine extracted sets of constraints. We also plan to conduct experiments with gold standard morphological lexicons on languages for which such resources exist \cite{Sylak:2015acl,Cotterell:2016acl}, and investigate approaches which learn morphological inflections and derivations in different languages automatically as another potential source of morphological constraints \cite[i.a.]{Soricut:2015naacl,Cotterell:2016sigmorphon,Faruqui:2016naacl,Kann:2017eacl,Aharoni:2017acl}.

\section{Downstream Task: Dialogue State Tracking (DST)}
\label{s:downstream}
Goal-oriented dialogue systems provide conversational interfaces for tasks such as booking flights or finding restaurants. In \emph{slot-based} systems, application domains are specified using \emph{ontologies} that define the search constraints which users can express. An ontology consists of a number of \emph{slots} and their assorted \emph{slot values}. In a \emph{restaurant search} domain, sets of slot-values could include \textsc{price} = [\emph{cheap, expensive}] or \textsc{food} = [\emph{Thai, Indian, ...}]. 

The DST model is the first component of modern dialogue pipelines \cite{young:10}. It serves to capture the intents expressed by the user at each dialogue turn and update the \emph{belief state}. This probability distribution over the possible dialogue states (defined by the domain ontology) is the system's internal estimate of the user's goals. It is used by the downstream \emph{dialogue manager} component to choose the subsequent system response \cite{su:2016:nnpolicy,Su:16}. The following example shows the true dialogue state in a multi-turn dialogue: \vspace{2.5mm}

{\footnotesize
\begin{tabular}{p{9cm}}
  \textbf{User:} What's good in the southern part of town?\\
  \texttt{inform(area=south)}\\[0.8ex]
  \textbf{System:} Vedanta is the top-rated Indian place. \\ [0.4ex]
  \textbf{User:} How about something cheaper? \\
  \texttt{inform(area=south, price=cheap)}\\[0.8ex]
  \textbf{System:} Seven Days is very popular. Great hot pot. \\ [0.4ex]
  \textbf{User:} What's the address?\\
  \texttt{inform(area=south, price=cheap); request(address)}\\[0.8ex]
  \textbf{System:} Seven Days is at 66 Regent Street. \\ [0.8ex]
\end{tabular}} \\

\noindent The Dialogue State Tracking Challenge (DSTC) shared task series formalised the evaluation and provided labelled DST datasets \cite{Henderson:14a, Henderson:14c,Williams:16}. While a plethora of DST models are available based on, e.g., hand-crafted rules \cite{Wang:2014} or conditional random fields \cite{Lee:13a}, the recent DST methodology has seen a shift towards neural-network architectures \cite[i.a.]{Henderson:14d,Henderson:14b,Zilka:15,Mrksic:15,Perez:16b,Liu:2017,Vodolan:2017,Mrksic:16b}.

\paragraph{Model: Neural Belief Tracker} 
To detect intents in user utterances, most existing models rely on either (or both): \textbf{1)} Spoken Language Understanding models which require large amounts of annotated training data; or \textbf{2)} hand-crafted, domain-specific lexicons which try to capture lexical and morphological variation. The Neural Belief Tracker (NBT) is a novel DST model which overcomes both issues by reasoning purely over pre-trained word vectors \cite{Mrksic:16b}. The NBT learns to compose these vectors into intermediate utterance and context representations. These are then used to decide which of the ontology-defined intents (goals) have been expressed by the user. The NBT model keeps word vectors \emph{fixed} during training, so that unseen, yet related words can be mapped to the right intent at test time (e.g.~\emph{northern} to \emph{north}).  


\paragraph{Data: Multilingual WOZ 2.0 Dataset} Our DST evaluation is based on the WOZ dataset, released by \newcite{Wen:17}. In this Wizard-of-Oz setup, two Amazon Mechanical Turk workers assumed the role of the user and the system asking/providing information about restaurants in Cambridge (operating over the same ontology and database used for DSTC2 \cite{Henderson:14a}). Users typed instead of speaking, removing the need to deal with noisy speech recognition. In DSTC datasets, users would quickly adapt to the system's inability to deal with complex queries. Conversely, the WOZ setup allowed them to use sophisticated language. The WOZ 2.0 release expanded the dataset to 1,200 dialogues \cite{Mrksic:16b}. In this work, we use translations of this dataset to Italian and German, released by \newcite{Mrksic:17}.       


\paragraph{Evaluation Setup} The principal metric we use to measure DST performance is  the \emph{joint goal accuracy}, which represents the proportion of test set dialogue turns where all user goals expressed up to that point of the dialogue were decoded correctly \cite{Henderson:14a}. The NBT models for \textsc{en}, \textsc{de} and \textsc{it} are trained using four variants of the \textsc{sgns-large} vectors: \textbf{1)} the initial distributional vectors; \textbf{2)} \emph{morph-fixed} vectors; \textbf{3)} and \textbf{4)} the two variants of \emph{morph-fitted} vectors (see Sect.~\ref{s:exp}). 

\begin{figure*}[t]
    \centering
    \begin{subfigure}[t]{0.45\linewidth}
        \centering
        \includegraphics[width=0.95\linewidth]{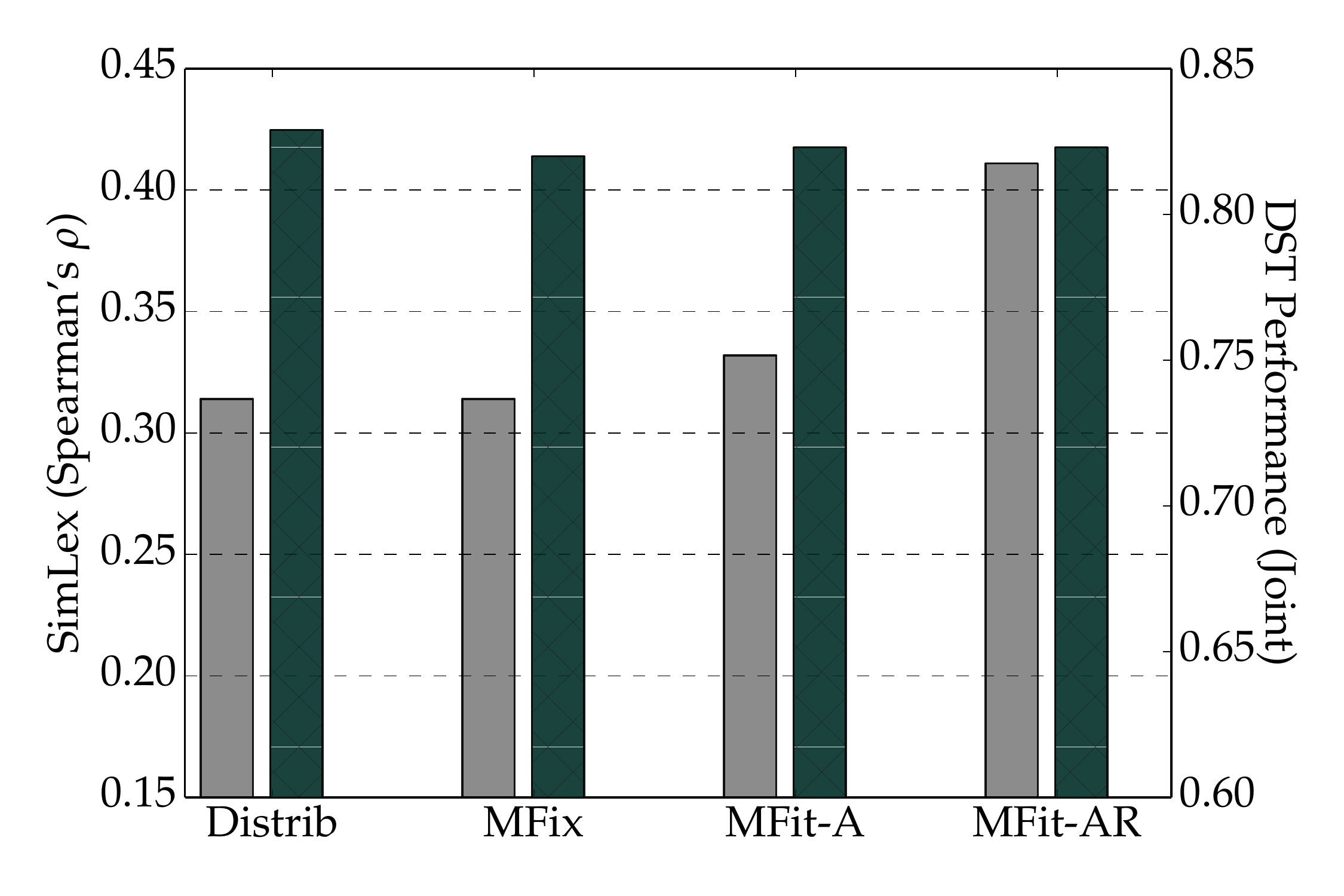}
        \caption{\textbf{English}}
        \label{fig:en}
    \end{subfigure}
    \begin{subfigure}[t]{0.45\textwidth}
        \centering
        \includegraphics[width=0.95\linewidth]{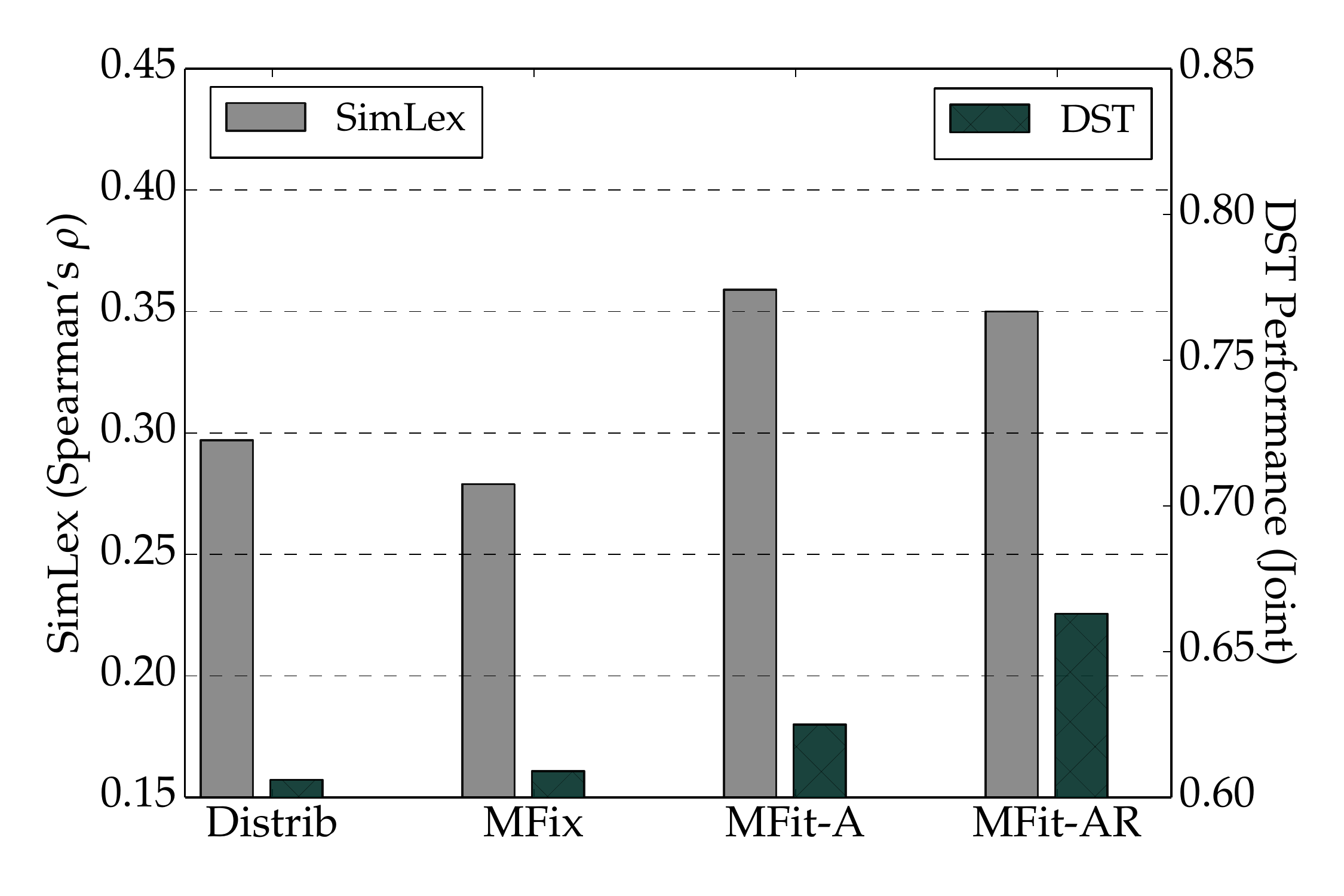}
        \caption{\textbf{German}}
        \label{fig:de}
    \end{subfigure}
    \begin{subfigure}[t]{0.45\linewidth}
        \centering
        \includegraphics[width=0.95\linewidth]{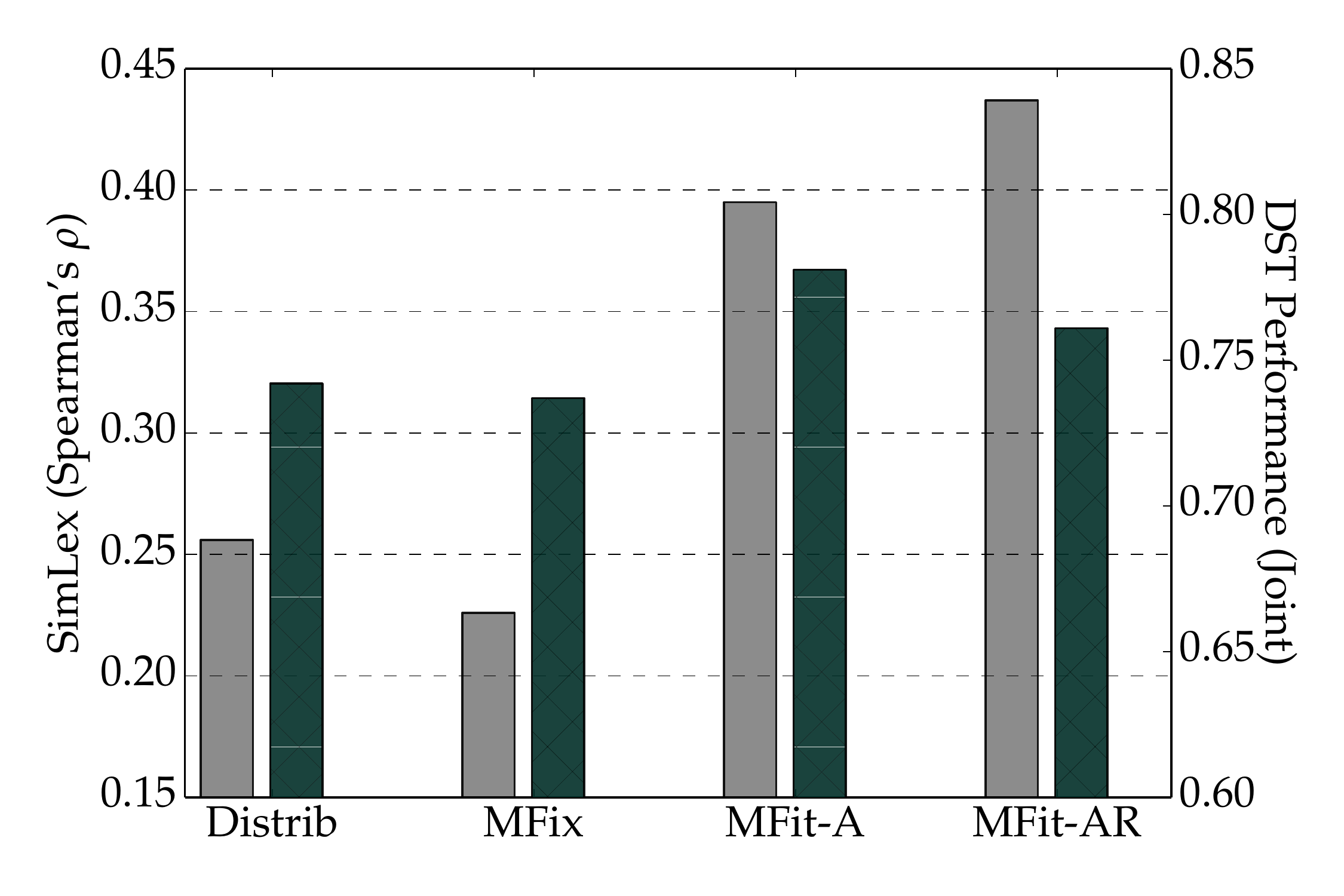}
        \caption{\textbf{Italian}}
        \label{fig:it}
    \end{subfigure}
    \begin{subfigure}[t]{0.43\textwidth}
        \centering
        \includegraphics[width=0.93\linewidth]{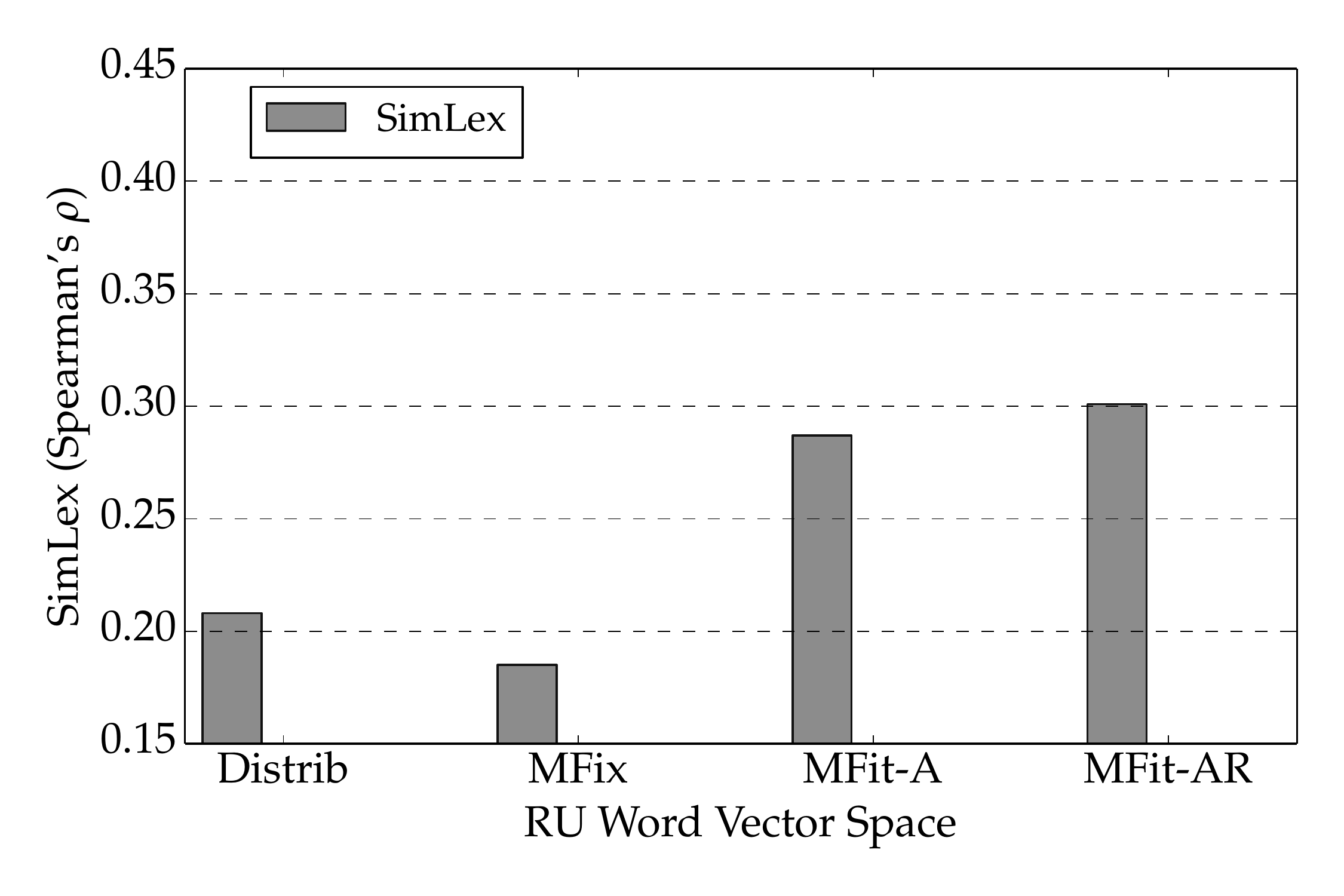}
        \caption{\textbf{Russian}}
        \label{fig:ru}
    \end{subfigure}
    \vspace{-0.5em}
    \caption{An overview of the results (Spearman's $\rho$ correlation) for four languages on SimLex-999 (grey bars, left $y$ axis) and the downstream DST performance (dark bars, right $y$ axis) using \textsc{sgns-large} vectors ($d=300$), see Tab.~\ref{tab:stats} and Sect.~\ref{s:exp}. The left $y$ axis measures the intrinsic word similarity performance, while the right $y$ axis provides the scale for the DST performance (there are no DST datasets for Russian).}
\vspace{-0.7mm}
\label{fig:main}
\end{figure*}

As shown by \newcite{Mrksic:17}, \emph{semantic specialisation} of the employed word vectors benefits DST performance across all three languages. However, large gains on SimLex-999 do not always induce correspondingly large gains in downstream performance. In our experiments, we investigate the extent to which \emph{morph-fitting} improves DST performance, and whether these gains exhibit stronger correlation with intrinsic performance. 

\paragraph{Results and Discussion} 
The dark bars (against the right axes) in Fig. \ref{fig:main} show the DST performance of NBT models making use of the four vector collections. \textsc{it} and \textsc{de} benefit from both kinds of \emph{morph-fitting}: \textsc{it} performance increases from $74.1 \rightarrow 78.1$ (\textsc{MFit-A}) and \textsc{de} performance rises even more: $60.6 \rightarrow 66.3$ (\textsc{MFit-AR}), setting a new state-of-the-art score for both datasets. The \emph{morph-fixed} vectors do not enhance DST performance, probably because fixing word vectors to their highest frequency inflectional form eliminates useful semantic content encoded in the original vectors. On the other hand, {morph-fitting} makes use of this information, supplementing it with semantic relations between different morphological forms. These conclusions are in line with the SimLex gains, where morph-fitting outperforms both distributional and \emph{morph-fixed} vectors. 


English performance shows little variation across the four word vector collections investigated here. This corroborates our intuition that, as a morphologically simpler language, English stands to gain less from fine-tuning the morphological variation for downstream applications. This result again points at the discrepancy between intrinsic and extrinsic evaluation: the considerable gains in SimLex performance do not necessarily induce similar gains in downstream performance. Additional discrepancies between SimLex and downstream DST performance are detected for German and Italian. While we observe a slight drop in SimLex performance with the \textsc{de} \textsc{MFit-AR} vectors compared to the \textsc{MFit-A} ones, their relative performance is reversed in the DST task. On the other hand, we see the opposite trend in Italian, where the \textsc{MFit-A} vectors score lower than the \textsc{MFit-AR} vectors on SimLex, but higher on the DST task. In summary, we believe these results show that SimLex is not a perfect proxy for downstream performance in language understanding tasks. Regardless, its performance does correlate with downstream performance to a large extent, providing a useful indicator for the usefulness of specific word vector spaces for extrinsic tasks such as DST. 

\section{Related Work}
\label{s:rw}
\paragraph{Semantic Specialisation} 
A standard approach to incorporating external information into vector spaces is to pull the representations of similar words closer together. Some models integrate such constraints into the training procedure, modifying the prior or the regularisation \cite{Yu:2014,Xu:2014,Bian:14,Kiela:2015emnlp}, or using a variant of the SGNS-style objective \cite{Liu:EtAl:15,Osborne:16}. Another class of models, popularly termed \textit{retrofitting}, injects lexical knowledge from available semantic databases (e.g., WordNet, PPDB) into pre-trained word vectors \cite{Faruqui:2015naacl,Jauhar:2015,Wieting:2015tacl,Nguyen:2016acl,Mrksic:2016naacl}. Morph-fitting falls into the latter category. However, instead of resorting to curated knowledge bases, and experimenting solely with English, we show that the \textit{morphological richness} of any language can be exploited as a source of inexpensive supervision for fine-tuning vector spaces, at the same time specialising them to better reflect true semantic similarity, and learning more accurate representations for low-frequency words.

\paragraph{Word Vectors and Morphology} 
The use of morphological resources to improve the representations of morphemes and words is an active area of research. The majority of proposed architectures encode morphological information, provided either as gold standard morphological resources \cite{Sylak:2015acl} such as CELEX \cite{Baayen:1995celex} or as an external analyser such as Morfessor \cite{Creutz:2007tslp}, along with distributional information jointly at \textit{training} time in the language modelling (LM) objective \cite[i.a.]{Luong:2013conll,Botha:2014icml,Qiu:2014coling,Cotterell:2015naacl,Bhatia:2016emnlp}. The key idea is to learn a morphological composition function \cite{Lazaridou:2013acl,Cotterell:2017tacl} which synthesises the representation of a word given the representations of its constituent morphemes. Contrary to our work, these models typically coalesce all lexical relations. 

Another class of models, operating at the character level, shares a similar methodology: such models compose token-level representations from subcomponent embeddings (subwords, morphemes, or characters) \cite[i.a.]{dosSantos:2014icml,Ling:2015emnlp,Cao:2016rep4,Kim:2016aaai,Wieting:2016emnlp,Verwimp:2017eacl}.

In contrast to prior work, our model \textit{decouples} the use of morphological information, now provided in the form of inflectional and derivational rules transformed into constraints, from the actual training. This pipelined approach results in a simpler, more portable model. In spirit, our work is similar to \newcite{Cotterell:2016acl}, who formulate the idea of post-training specialisation in a generative Bayesian framework. Their work uses gold morphological lexicons; we show that competitive performance can be achieved using a non-exhaustive set of simple rules. Our framework facilitates the inclusion of \textit{antonyms} at no extra cost and naturally extends to constraints from other sources (e.g., WordNet) in future work. Another practical difference is that we focus on similarity and evaluate {morph-fitting} in a well-defined downstream task where the artefacts of the distributional hypothesis are known to prompt statistical system failures. 

\section{Conclusion and Future Work}
\label{s:conclusion}
We have presented a novel \textit{morph-fitting} method which injects morphological knowledge in the form of linguistic constraints into word vector spaces. The method makes use of implicit semantic signals encoded in inflectional and derivational rules which describe the morphological processes in a language. The results in intrinsic word similarity tasks show that \textit{morph-fitting} improves vector spaces induced by distributional models across four languages. Finally, we have shown that the use of \emph{morph-fitted} vectors boosts the performance of downstream language understanding models which rely on word representations as features, especially for morphologically rich languages such as German.

Future work will focus on other potential sources of morphological knowledge, porting the framework to other morphologically rich languages and downstream tasks, and on further refinements of the post-processing specialisation algorithm and the constraint selection.

\section*{Acknowledgments}
This work is supported by the ERC Consolidator Grant LEXICAL: Lexical Acquisition Across Languages (no 648909). RR is supported by the Intel-ICRI grant: Hybrid Models for Minimally Supervised Information Extraction from Conversations. The authors are grateful to the anonymous reviewers for their helpful suggestions.



\bibliography{acl2017_refs}

\begin{thebibliography}{}
\expandafter\ifx\csname natexlab\endcsname\relax\def\natexlab#1{#1}\fi

\bibitem[{Aharoni and Goldberg(2017)}]{Aharoni:2017acl}
Roee Aharoni and Yoav Goldberg. 2017.
\newblock \href{https://arxiv.org/abs/1611.01487}{Morphological inflection
  generation with hard monotonic attention}.
\newblock In {\em Proceedings of ACL\/}.
\newblock
  \href{https://arxiv.org/abs/1611.01487}{https://arxiv.org/abs/1611.01487}.

\bibitem[{Al-Rfou et~al.(2013)Al-Rfou, Perozzi, and Skiena}]{AlRfou:2013conll}
Rami Al-Rfou, Bryan Perozzi, and Steven Skiena. 2013.
\newblock \href{http://www.aclweb.org/anthology/W13-3520}{Polyglot:
  {D}istributed word representations for multilingual {NLP}}.
\newblock In {\em Proceedings of CoNLL\/}. pages 183--192.
\newblock
  \href{http://www.aclweb.org/anthology/W13-3520}{http://www.aclweb.org/anthology/W13-3520}.

\bibitem[{Avramidis and Koehn(2008)}]{Avramidis:2008acl}
Eleftherios Avramidis and Philipp Koehn. 2008.
\newblock \href{http://www.aclweb.org/anthology/P/P08/P08-1087}{Enriching
  morphologically poor languages for statistical machine translation}.
\newblock In {\em Proceedings of ACL\/}. pages 763--770.
\newblock
  \href{http://www.aclweb.org/anthology/P/P08/P08-1087}{http://www.aclweb.org/anthology/P/P08/P08-1087}.

\bibitem[{Baayen et~al.(1995)Baayen, Piepenbrock, and van
  Rijn}]{Baayen:1995celex}
Harald~R. Baayen, Richard Piepenbrock, and Hedderik van Rijn. 1995.
\newblock The {CELEX lexical data base on CD-ROM} .

\bibitem[{Bansal et~al.(2014)Bansal, Gimpel, and Livescu}]{Bansal:2014acl}
Mohit Bansal, Kevin Gimpel, and Karen Livescu. 2014.
\newblock \href{http://www.aclweb.org/anthology/P14-2131}{Tailoring continuous
  word representations for dependency parsing}.
\newblock In {\em Proceedings of ACL\/}. pages 809--815.
\newblock
  \href{http://www.aclweb.org/anthology/P14-2131}{http://www.aclweb.org/anthology/P14-2131}.

\bibitem[{Baroni et~al.(2014)Baroni, Dinu, and Kruszewski}]{Baroni:2014acl}
Marco Baroni, Georgiana Dinu, and Germ{\'{a}}n Kruszewski. 2014.
\newblock \href{http://www.aclweb.org/anthology/P14-1023}{Don't count, predict!
  {A} systematic comparison of context-counting vs. context-predicting semantic
  vectors}.
\newblock In {\em Proceedings of ACL\/}. pages 238--247.
\newblock
  \href{http://www.aclweb.org/anthology/P14-1023}{http://www.aclweb.org/anthology/P14-1023}.

\bibitem[{Bhatia et~al.(2016)Bhatia, Guthrie, and
  Eisenstein}]{Bhatia:2016emnlp}
Parminder Bhatia, Robert Guthrie, and Jacob Eisenstein. 2016.
\newblock \href{https://aclweb.org/anthology/D16-1047}{Morphological priors for
  probabilistic neural word embeddings}.
\newblock In {\em Proceedings of EMNLP\/}. pages 490--500.
\newblock
  \href{https://aclweb.org/anthology/D16-1047}{https://aclweb.org/anthology/D16-1047}.

\bibitem[{Bian et~al.(2014)Bian, Gao, and Liu}]{Bian:14}
Jiang Bian, Bin Gao, and Tie{-}Yan Liu. 2014.
\newblock \href{https://doi.org/10.1007/978-3-662-44848-9\_9}{Knowledge-powered
  deep learning for word embedding}.
\newblock In {\em Proceedings of ECML-PKDD\/}. pages 132--148.
\newblock
  \href{https://doi.org/10.1007/978-3-662-44848-9\_9}{https://doi.org/10.1007/978-3-662-44848-9\_9}.

\bibitem[{Botha and Blunsom(2014)}]{Botha:2014icml}
Jan~A. Botha and Phil Blunsom. 2014.
\newblock
  \href{http://jmlr.org/proceedings/papers/v32/botha14.html}{Compositional
  morphology for word representations and language modelling}.
\newblock In {\em Proceedings of ICML\/}. pages 1899--1907.
\newblock
  \href{http://jmlr.org/proceedings/papers/v32/botha14.html}{http://jmlr.org/proceedings/papers/v32/botha14.html}.

\bibitem[{Bruni et~al.(2014)Bruni, Tran, and Baroni}]{Bruni:2014jair}
Elia Bruni, Nam{-}Khanh Tran, and Marco Baroni. 2014.
\newblock \href{https://doi.org/10.1613/jair.4135}{Multimodal distributional
  semantics}.
\newblock {\em Journal of Artificial Intelligence Research\/} 49:1--47.
\newblock
  \href{https://doi.org/10.1613/jair.4135}{https://doi.org/10.1613/jair.4135}.

\bibitem[{Cao and Rei(2016)}]{Cao:2016rep4}
Kris Cao and Marek Rei. 2016.
\newblock \href{http://aclweb.org/anthology/W/W16/W16-1603}{A joint model for
  word embedding and word morphology}.
\newblock In {\em Proceedings of the 1st Workshop on Representation Learning
  for NLP\/}. pages 18--26.
\newblock
  \href{http://aclweb.org/anthology/W/W16/W16-1603}{http://aclweb.org/anthology/W/W16/W16-1603}.

\bibitem[{Chen and Manning(2014)}]{Chen:2014emnlp}
Danqi Chen and Christopher~D. Manning. 2014.
\newblock \href{http://www.aclweb.org/anthology/D14-1082}{A fast and accurate
  dependency parser using neural networks}.
\newblock In {\em Proceedings of EMNLP\/}. pages 740--750.
\newblock
  \href{http://www.aclweb.org/anthology/D14-1082}{http://www.aclweb.org/anthology/D14-1082}.

\bibitem[{Collobert et~al.(2011)Collobert, Weston, Bottou, Karlen, Kavukcuoglu,
  and Kuksa}]{Collobert:2011jmlr}
Ronan Collobert, Jason Weston, L{\'{e}}on Bottou, Michael Karlen, Koray
  Kavukcuoglu, and Pavel~P. Kuksa. 2011.
\newblock \href{http://dl.acm.org/citation.cfm?id=1953048.2078186}{Natural
  language processing (almost) from scratch}.
\newblock {\em Journal of Machine Learning Research\/} 12:2493--2537.
\newblock
  \href{http://dl.acm.org/citation.cfm?id=1953048.2078186}{http://dl.acm.org/citation.cfm?id=1953048.2078186}.

\bibitem[{Cotterell et~al.(2016{\natexlab{a}})Cotterell, Kirov, Sylak-Glassman,
  Yarowsky, Eisner, and Hulden}]{Cotterell:2016sigmorphon}
Ryan Cotterell, Christo Kirov, John Sylak-Glassman, David Yarowsky, Jason
  Eisner, and Mans Hulden. 2016{\natexlab{a}}.
\newblock \href{http://anthology.aclweb.org/W16-2002}{The sigmorphon 2016
  shared task - morphological reinflection}.
\newblock In {\em Proceedings of the 14th SIGMORPHON Workshop on Computational
  Research in Phonetics, Phonology, and Morphology\/}. pages 10--22.
\newblock
  \href{http://anthology.aclweb.org/W16-2002}{http://anthology.aclweb.org/W16-2002}.

\bibitem[{Cotterell and Sch\"{u}tze(2015)}]{Cotterell:2015naacl}
Ryan Cotterell and Hinrich Sch\"{u}tze. 2015.
\newblock \href{http://www.aclweb.org/anthology/N15-1140}{Morphological
  word-embeddings}.
\newblock In {\em Proceedings of NAACL-HLT\/}. pages 1287--1292.
\newblock
  \href{http://www.aclweb.org/anthology/N15-1140}{http://www.aclweb.org/anthology/N15-1140}.

\bibitem[{Cotterell and Sch\"{u}tze(2017)}]{Cotterell:2017tacl}
Ryan Cotterell and Hinrich Sch\"{u}tze. 2017.
\newblock \href{https://arxiv.org/abs/1701.00946}{Joint semantic synthesis and
  morphological analysis of the derived word}.
\newblock {\em Transactions of the ACL\/}
  \href{https://arxiv.org/abs/1701.00946}{https://arxiv.org/abs/1701.00946}.

\bibitem[{Cotterell et~al.(2016{\natexlab{b}})Cotterell, Sch\"{u}tze, and
  Eisner}]{Cotterell:2016acl}
Ryan Cotterell, Hinrich Sch\"{u}tze, and Jason Eisner. 2016{\natexlab{b}}.
\newblock \href{http://www.aclweb.org/anthology/P16-1156}{Morphological
  smoothing and extrapolation of word embeddings}.
\newblock In {\em Proceedings of ACL\/}. pages 1651--1660.
\newblock
  \href{http://www.aclweb.org/anthology/P16-1156}{http://www.aclweb.org/anthology/P16-1156}.

\bibitem[{Creutz and Lagus(2007)}]{Creutz:2007tslp}
Mathias Creutz and Krista Lagus. 2007.
\newblock \href{http://doi.acm.org/10.1145/1217098.1217101}{Unsupervised models
  for morpheme segmentation and morphology learning}.
\newblock {\em {TSLP}\/} 4(1):3:1--3:34.
\newblock
  \href{http://doi.acm.org/10.1145/1217098.1217101}{http://doi.acm.org/10.1145/1217098.1217101}.

\bibitem[{Curran(2004)}]{Curran:04}
James Curran. 2004.
\newblock {\em From Distributional to Semantic Similarity\/}.
\newblock Ph.D. thesis, School of Informatics, University of Edinburgh.
\newblock
  \href{http://hdl.handle.net/1842/563}{http://hdl.handle.net/1842/563}.

\bibitem[{Dinu et~al.(2015)Dinu, Lazaridou, and Baroni}]{Dinu:2015arxiv}
Georgiana Dinu, Angeliki Lazaridou, and Marco Baroni. 2015.
\newblock \href{http://arxiv.org/abs/1412.6568}{Improving zero-shot learning by
  mitigating the hubness problem}.
\newblock In {\em Proceedings of ICLR (Workshop Papers)\/}.
\newblock
  \href{http://arxiv.org/abs/1412.6568}{http://arxiv.org/abs/1412.6568}.

\bibitem[{dos Santos and Zadrozny(2014)}]{dosSantos:2014icml}
C{\'{\i}}cero~Nogueira dos Santos and Bianca Zadrozny. 2014.
\newblock \href{http://jmlr.org/proceedings/papers/v32/santos14.html}{Learning
  character-level representations for part-of-speech tagging}.
\newblock In {\em Proceedings of ICML\/}. pages 1818--1826.
\newblock
  \href{http://jmlr.org/proceedings/papers/v32/santos14.html}{http://jmlr.org/proceedings/papers/v32/santos14.html}.

\bibitem[{Duchi et~al.(2011)Duchi, Hazan, and Singer}]{Duchi:11}
John~C. Duchi, Elad Hazan, and Yoram Singer. 2011.
\newblock \href{http://dl.acm.org/citation.cfm?id=2021068}{Adaptive subgradient
  methods for online learning and stochastic optimization}.
\newblock {\em Journal of Machine Learning Research\/} 12:2121--2159.
\newblock
  \href{http://dl.acm.org/citation.cfm?id=2021068}{http://dl.acm.org/citation.cfm?id=2021068}.

\bibitem[{Ehrmann et~al.(2014)Ehrmann, Cecconi, Vannella, Mccrae, Cimiano, and
  Navigli}]{Ehrmann:14}
Maud Ehrmann, Francesco Cecconi, Daniele Vannella, John~Philip Mccrae, Philipp
  Cimiano, and Roberto Navigli. 2014.
\newblock
  \href{http://www.lrec-conf.org/proceedings/lrec2014/summaries/810.html}{Representing
  multilingual data as linked data: {T}he case of {BabelNet 2.0}}.
\newblock In {\em Proceedings of LREC\/}. pages 401--408.
\newblock
  \href{http://www.lrec-conf.org/proceedings/lrec2014/summaries/810.html}{http://www.lrec-conf.org/proceedings/lrec2014/summaries/810.html}.

\bibitem[{Faruqui et~al.(2015)Faruqui, Dodge, Jauhar, Dyer, Hovy, and
  Smith}]{Faruqui:2015naacl}
Manaal Faruqui, Jesse Dodge, Sujay~Kumar Jauhar, Chris Dyer, Eduard Hovy, and
  Noah~A. Smith. 2015.
\newblock \href{http://www.aclweb.org/anthology/N15-1184}{Retrofitting word
  vectors to semantic lexicons}.
\newblock In {\em Proceedings of NAACL-HLT\/}. pages 1606--1615.
\newblock
  \href{http://www.aclweb.org/anthology/N15-1184}{http://www.aclweb.org/anthology/N15-1184}.

\bibitem[{Faruqui and Dyer(2014)}]{Faruqui:2014eacl}
Manaal Faruqui and Chris Dyer. 2014.
\newblock \href{http://www.aclweb.org/anthology/E14-1049}{Improving vector
  space word representations using multilingual correlation}.
\newblock In {\em Proceedings of EACL\/}. pages 462--471.
\newblock
  \href{http://www.aclweb.org/anthology/E14-1049}{http://www.aclweb.org/anthology/E14-1049}.

\bibitem[{Faruqui et~al.(2016)Faruqui, Tsvetkov, Neubig, and
  Dyer}]{Faruqui:2016naacl}
Manaal Faruqui, Yulia Tsvetkov, Graham Neubig, and Chris Dyer. 2016.
\newblock \href{http://www.aclweb.org/anthology/N16-1077}{Morphological
  inflection generation using character sequence to sequence learning}.
\newblock In {\em Proceedings of NAACL-HLT\/}. pages 634--643.
\newblock
  \href{http://www.aclweb.org/anthology/N16-1077}{http://www.aclweb.org/anthology/N16-1077}.

\bibitem[{Fellbaum(1998)}]{Fellbaum:1998wn}
Christiane Fellbaum. 1998.
\newblock {\em WordNet\/}.
\newblock
  \href{https://mitpress.mit.edu/books/wordnet}{https://mitpress.mit.edu/books/wordnet}.

\bibitem[{Finkelstein et~al.(2002)Finkelstein, Gabrilovich, Matias, Rivlin,
  Solan, Wolfman, and Ruppin}]{Finkelstein:2002tois}
Lev Finkelstein, Evgeniy Gabrilovich, Yossi Matias, Ehud Rivlin, Zach Solan,
  Gadi Wolfman, and Eytan Ruppin. 2002.
\newblock \href{https://doi.org/10.1145/503104.503110}{Placing search in
  context: {T}he concept revisited}.
\newblock {\em {ACM} Transactions on Information Systems\/} 20(1):116--131.
\newblock
  \href{https://doi.org/10.1145/503104.503110}{https://doi.org/10.1145/503104.503110}.

\bibitem[{Fromkin et~al.(2013)Fromkin, Rodman, and Hyams}]{Fromkin:2013book}
Victoria Fromkin, Robert Rodman, and Nina Hyams. 2013.
\newblock {\em An Introduction to Language, 10th Edition\/}.

\bibitem[{Ganitkevitch et~al.(2013)Ganitkevitch, Van~Durme, and
  Callison-Burch}]{Ganitkevitch:2013naacl}
Juri Ganitkevitch, Benjamin Van~Durme, and Chris Callison-Burch. 2013.
\newblock \href{http://www.aclweb.org/anthology/N13-1092}{{PPDB: The Paraphrase
  Database}}.
\newblock In {\em Proceedings of NAACL-HLT\/}. pages 758--764.
\newblock
  \href{http://www.aclweb.org/anthology/N13-1092}{http://www.aclweb.org/anthology/N13-1092}.

\bibitem[{Gerz et~al.(2016)Gerz, Vuli{\'{c}}, Hill, Reichart, and
  Korhonen}]{Gerz:2016emnlp}
Daniela Gerz, Ivan Vuli{\'{c}}, Felix Hill, Roi Reichart, and Anna Korhonen.
  2016.
\newblock \href{https://aclweb.org/anthology/D16-1235}{{SimVerb-3500:} {A}
  large-scale evaluation set of verb similarity}.
\newblock In {\em Proceedings of EMNLP\/}. pages 2173--2182.
\newblock
  \href{https://aclweb.org/anthology/D16-1235}{https://aclweb.org/anthology/D16-1235}.

\bibitem[{Harris(1954)}]{Harris:1954}
Zellig~S. Harris. 1954.
\newblock Distributional structure.
\newblock {\em Word\/} 10(23):146--162.

\bibitem[{Haspelmath and Sims(2013)}]{Haspelmath:2013book}
Martin Haspelmath and Andrea Sims. 2013.
\newblock {\em Understanding morphology\/}.

\bibitem[{Henderson et~al.(2014{\natexlab{a}})Henderson, Thomson, and
  Wiliams}]{Henderson:14a}
Matthew Henderson, Blaise Thomson, and Jason~D. Wiliams. 2014{\natexlab{a}}.
\newblock \href{http://aclweb.org/anthology/W/W14/W14-4337.pdf}{The {Second
  Dialog State Tracking Challenge}}.
\newblock In {\em Proceedings of SIGDIAL\/}. pages 263--272.
\newblock
  \href{http://aclweb.org/anthology/W/W14/W14-4337.pdf}{http://aclweb.org/anthology/W/W14/W14-4337.pdf}.

\bibitem[{Henderson et~al.(2014{\natexlab{b}})Henderson, Thomson, and
  Wiliams}]{Henderson:14c}
Matthew Henderson, Blaise Thomson, and Jason~D. Wiliams. 2014{\natexlab{b}}.
\newblock \href{https://doi.org/10.1109/SLT.2014.7078595}{The {Third Dialog
  State Tracking Challenge}}.
\newblock In {\em Proceedings of IEEE SLT\/}. pages 324--329.
\newblock
  \href{https://doi.org/10.1109/SLT.2014.7078595}{https://doi.org/10.1109/SLT.2014.7078595}.

\bibitem[{Henderson et~al.(2014{\natexlab{c}})Henderson, Thomson, and
  Young}]{Henderson:14d}
Matthew Henderson, Blaise Thomson, and Steve Young. 2014{\natexlab{c}}.
\newblock Robust dialog state tracking using delexicalised recurrent neural
  networks and unsupervised adaptation.
\newblock In {\em Proceedings of IEEE SLT\/}. pages 360--365.

\bibitem[{Henderson et~al.(2014{\natexlab{d}})Henderson, Thomson, and
  Young}]{Henderson:14b}
Matthew Henderson, Blaise Thomson, and Steve Young. 2014{\natexlab{d}}.
\newblock \href{http://aclweb.org/anthology/W/W14/W14-4340.pdf}{Word-based
  dialog state tracking with recurrent neural networks}.
\newblock In {\em Proceedings of SIGDIAL\/}. pages 292--299.
\newblock
  \href{http://aclweb.org/anthology/W/W14/W14-4340.pdf}{http://aclweb.org/anthology/W/W14/W14-4340.pdf}.

\bibitem[{Hill et~al.(2015)Hill, Reichart, and Korhonen}]{Hill:2015cl}
Felix Hill, Roi Reichart, and Anna Korhonen. 2015.
\newblock \href{https://doi.org/10.1162/COLI\_a\_00237}{{SimLex-999:
  E}valuating semantic models with (genuine) similarity estimation}.
\newblock {\em Computational Linguistics\/} 41(4):665--695.
\newblock
  \href{https://doi.org/10.1162/COLI\_a\_00237}{https://doi.org/10.1162/COLI\_a\_00237}.

\bibitem[{Jauhar et~al.(2015)Jauhar, Dyer, and Hovy}]{Jauhar:2015}
Sujay~Kumar Jauhar, Chris Dyer, and Eduard~H. Hovy. 2015.
\newblock \href{http://www.aclweb.org/anthology/N15-1070}{Ontologically
  grounded multi-sense representation learning for semantic vector space
  models}.
\newblock In {\em Proceedings of NAACL\/}. pages 683--693.
\newblock
  \href{http://www.aclweb.org/anthology/N15-1070}{http://www.aclweb.org/anthology/N15-1070}.

\bibitem[{Johannsen et~al.(2015)Johannsen, Mart\'{i}nez~Alonso, and
  S{\o}gaard}]{Johannsen:2015emnlp}
Anders Johannsen, H\'{e}ctor Mart\'{i}nez~Alonso, and Anders S{\o}gaard. 2015.
\newblock \href{http://aclweb.org/anthology/D15-1245}{Any-language
  frame-semantic parsing}.
\newblock In {\em Proceedings of EMNLP\/}. pages 2062--2066.
\newblock
  \href{http://aclweb.org/anthology/D15-1245}{http://aclweb.org/anthology/D15-1245}.

\bibitem[{Kann et~al.(2017)Kann, Cotterell, and Sch\"{u}tze}]{Kann:2017eacl}
Katharina Kann, Ryan Cotterell, and Hinrich Sch\"{u}tze. 2017.
\newblock \href{http://www.aclweb.org/anthology/E17-1049}{Neural multi-source
  morphological reinflection}.
\newblock In {\em Proceedings of EACL\/}. pages 514--524.
\newblock
  \href{http://www.aclweb.org/anthology/E17-1049}{http://www.aclweb.org/anthology/E17-1049}.

\bibitem[{Kiela et~al.(2015)Kiela, Hill, and Clark}]{Kiela:2015emnlp}
Douwe Kiela, Felix Hill, and Stephen Clark. 2015.
\newblock \href{http://aclweb.org/anthology/D15-1242}{Specializing word
  embeddings for similarity or relatedness}.
\newblock In {\em Proceedings of EMNLP\/}. pages 2044--2048.
\newblock
  \href{http://aclweb.org/anthology/D15-1242}{http://aclweb.org/anthology/D15-1242}.

\bibitem[{Kim et~al.(2016)Kim, Jernite, Sontag, and Rush}]{Kim:2016aaai}
Yoon Kim, Yacine Jernite, David Sontag, and Alexander~M. Rush. 2016.
\newblock Character-aware neural language models.
\newblock In {\em Proceedings of AAAI\/}. pages 2741--2749.

\bibitem[{Lazaridou et~al.(2013)Lazaridou, Marelli, Zamparelli, and
  Baroni}]{Lazaridou:2013acl}
Angeliki Lazaridou, Marco Marelli, Roberto Zamparelli, and Marco Baroni. 2013.
\newblock \href{http://www.aclweb.org/anthology/P13-1149}{Compositional-ly
  derived representations of morphologically complex words in distributional
  semantics}.
\newblock In {\em Proceedings of ACL\/}. pages 1517--1526.
\newblock
  \href{http://www.aclweb.org/anthology/P13-1149}{http://www.aclweb.org/anthology/P13-1149}.

\bibitem[{Lee and Eskenazi(2013)}]{Lee:13a}
Sungjin Lee and Maxine Eskenazi. 2013.
\newblock \href{http://aclweb.org/anthology/W/W13/W13-4066.pdf}{Recipe for
  building robust spoken dialog state trackers: {Dialog State Tracking
  Challenge} system description}.
\newblock In {\em Proceedings of SIGDIAL\/}. pages 414--422.
\newblock
  \href{http://aclweb.org/anthology/W/W13/W13-4066.pdf}{http://aclweb.org/anthology/W/W13/W13-4066.pdf}.

\bibitem[{Leviant and Reichart(2015)}]{Leviant:2015arxiv}
Ira Leviant and Roi Reichart. 2015.
\newblock \href{http://arxiv.org/abs/1508.00106}{Separated by an un-common
  language: {T}owards judgment language informed vector space modeling}.
\newblock {\em CoRR\/} abs/1508.00106.
\newblock
  \href{http://arxiv.org/abs/1508.00106}{http://arxiv.org/abs/1508.00106}.

\bibitem[{Levy and Goldberg(2014)}]{Levy:2014acl}
Omer Levy and Yoav Goldberg. 2014.
\newblock \href{http://www.aclweb.org/anthology/P14-2050}{Dependency-based word
  embeddings}.
\newblock In {\em Proceedings of ACL\/}. pages 302--308.
\newblock
  \href{http://www.aclweb.org/anthology/P14-2050}{http://www.aclweb.org/anthology/P14-2050}.

\bibitem[{Ling et~al.(2015)Ling, Dyer, Black, Trancoso, Fermandez, Amir,
  Marujo, and Luis}]{Ling:2015emnlp}
Wang Ling, Chris Dyer, Alan~W. Black, Isabel Trancoso, Ramon Fermandez, Silvio
  Amir, Luis Marujo, and Tiago Luis. 2015.
\newblock \href{http://aclweb.org/anthology/D15-1176}{Finding function in form:
  {C}ompositional character models for open vocabulary word representation}.
\newblock In {\em Proceedings of EMNLP\/}. pages 1520--1530.
\newblock
  \href{http://aclweb.org/anthology/D15-1176}{http://aclweb.org/anthology/D15-1176}.

\bibitem[{Liu and Perez(2017)}]{Liu:2017}
Fei Liu and Julien Perez. 2017.
\newblock \href{http://www.aclweb.org/anthology/E17-1001}{Gated end-to-end
  memory networks}.
\newblock In {\em Proceedings of EACL\/}. pages 1--10.
\newblock
  \href{http://www.aclweb.org/anthology/E17-1001}{http://www.aclweb.org/anthology/E17-1001}.

\bibitem[{Liu et~al.(2015)Liu, Jiang, Wei, Ling, and Hu}]{Liu:EtAl:15}
Quan Liu, Hui Jiang, Si~Wei, Zhen-Hua Ling, and Yu~Hu. 2015.
\newblock \href{http://www.aclweb.org/anthology/P15-1145}{Learning semantic
  word embeddings based on ordinal knowledge constraints}.
\newblock In {\em Proceedings of ACL\/}. pages 1501--1511.
\newblock
  \href{http://www.aclweb.org/anthology/P15-1145}{http://www.aclweb.org/anthology/P15-1145}.

\bibitem[{Luong et~al.(2015)Luong, Pham, and Manning}]{Luong:2015naacl}
Thang Luong, Hieu Pham, and Christopher~D. Manning. 2015.
\newblock \href{http://www.aclweb.org/anthology/W15-1521}{Bilingual word
  representations with monolingual quality in mind}.
\newblock In {\em Proceedings of the 1st Workshop on Vector Space Modeling for
  Natural Language Processing\/}. pages 151--159.
\newblock
  \href{http://www.aclweb.org/anthology/W15-1521}{http://www.aclweb.org/anthology/W15-1521}.

\bibitem[{Luong et~al.(2013)Luong, Socher, and Manning}]{Luong:2013conll}
Thang Luong, Richard Socher, and Christopher Manning. 2013.
\newblock \href{http://www.aclweb.org/anthology/W13-3512}{Better word
  representations with recursive neural networks for morphology}.
\newblock In {\em Proceedings of CoNLL\/}. pages 104--113.
\newblock
  \href{http://www.aclweb.org/anthology/W13-3512}{http://www.aclweb.org/anthology/W13-3512}.

\bibitem[{Melamud et~al.(2016)Melamud, Goldberger, and
  Dagan}]{Melamud:2016conll}
Oren Melamud, Jacob Goldberger, and Ido Dagan. 2016.
\newblock \href{http://aclweb.org/anthology/K/K16/K16-1006.pdf}{{Context2vec:
  L}earning generic context embedding with bidirectional {LSTM}}.
\newblock In {\em Proceedings of CoNLL\/}. pages 51--61.
\newblock
  \href{http://aclweb.org/anthology/K/K16/K16-1006.pdf}{http://aclweb.org/anthology/K/K16/K16-1006.pdf}.

\bibitem[{Mikolov et~al.(2013)Mikolov, Sutskever, Chen, Corrado, and
  Dean}]{Mikolov:2013nips}
Tomas Mikolov, Ilya Sutskever, Kai Chen, Gregory~S. Corrado, and Jeffrey Dean.
  2013.
\newblock Distributed representations of words and phrases and their
  compositionality.
\newblock In {\em Proceedings of NIPS\/}. pages 3111--3119.

\bibitem[{Mnih and Kavukcuoglu(2013)}]{Mnih:2013nips}
Andriy Mnih and Koray Kavukcuoglu. 2013.
\newblock Learning word embeddings efficiently with noise-contrastive
  estimation.
\newblock In {\em Proceedings of NIPS\/}. pages 2265--2273.

\bibitem[{Mrk\v{s}i\'c et~al.(2015)Mrk\v{s}i\'c, {\'O S\'eaghdha}, Thomson,
  Ga\v{s}i\'c, Su, Vandyke, Wen, and Young}]{Mrksic:15}
Nikola Mrk\v{s}i\'c, Diarmuid {\'O S\'eaghdha}, Blaise Thomson, Milica
  Ga\v{s}i\'c, Pei-Hao Su, David Vandyke, Tsung-Hsien Wen, and Steve Young.
  2015.
\newblock \href{http://aclweb.org/anthology/P/P15/P15-2130.pdf}{Multi-domain
  dialog state tracking using recurrent neural networks}.
\newblock In {\em Proceedings of ACL\/}. pages 794--799.
\newblock
  \href{http://aclweb.org/anthology/P/P15/P15-2130.pdf}{http://aclweb.org/anthology/P/P15/P15-2130.pdf}.

\bibitem[{Mrk\v{s}i\'c et~al.(2017{\natexlab{a}})Mrk\v{s}i\'c, {\'O
  S\'eaghdha}, Thomson, Wen, and Young}]{Mrksic:16b}
Nikola Mrk\v{s}i\'c, Diarmuid {\'O S\'eaghdha}, Blaise Thomson, Tsung-Hsien
  Wen, and Steve Young. 2017{\natexlab{a}}.
\newblock \href{http://arxiv.org/abs/1606.03777}{{Neural Belief Tracker:
  D}ata-driven dialogue state tracking}.
\newblock In {\em Proceedings of ACL\/}.
\newblock
  \href{http://arxiv.org/abs/1606.03777}{http://arxiv.org/abs/1606.03777}.

\bibitem[{Mrk\v{s}i\'{c} et~al.(2016)Mrk\v{s}i\'{c}, S{\'{e}}aghdha, Thomson,
  Ga\v{s}i\'{c}, Rojas{-}Barahona, Su, Vandyke, Wen, and
  Young}]{Mrksic:2016naacl}
Nikola Mrk\v{s}i\'{c}, Diarmuid~{\'{O}} S{\'{e}}aghdha, Blaise Thomson, Milica
  Ga\v{s}i\'{c}, Lina~Maria Rojas{-}Barahona, Pei{-}Hao Su, David Vandyke,
  Tsung{-}Hsien Wen, and Steve Young. 2016.
\newblock \href{http://aclweb.org/anthology/N/N16/N16-1018.pdf}{Counter-fitting
  word vectors to linguistic constraints}.
\newblock In {\em Proceedings of NAACL-HLT\/}.
\newblock
  \href{http://aclweb.org/anthology/N/N16/N16-1018.pdf}{http://aclweb.org/anthology/N/N16/N16-1018.pdf}.

\bibitem[{Mrk\v{s}i\'c et~al.(2017{\natexlab{b}})Mrk\v{s}i\'c, Vuli\'{c}, {\'O
  S\'eaghdha}, Reichart, Ga\v{s}i\'{c}, Korhonen, and Young}]{Mrksic:17}
Nikola Mrk\v{s}i\'c, Ivan Vuli\'{c}, Diarmuid {\'O S\'eaghdha}, Roi Reichart,
  Milica Ga\v{s}i\'{c}, Anna Korhonen, and Steve Young. 2017{\natexlab{b}}.
\newblock {Semantic Specialisation of Distributional Word Vector Spaces using
  Monolingual and Cross-Lingual Constraints}.
\newblock arXiv.

\bibitem[{Nair and Hinton(2010)}]{Nair:2010icml}
Vinod Nair and Geoffrey~E. Hinton. 2010.
\newblock \href{http://www.icml2010.org/papers/432.pdf}{Rectified linear units
  improve restricted {B}oltzmann machines}.
\newblock In {\em Proceedings of ICML\/}. pages 807--814.
\newblock
  \href{http://www.icml2010.org/papers/432.pdf}{http://www.icml2010.org/papers/432.pdf}.

\bibitem[{Navigli and Ponzetto(2012)}]{Navigli:12}
Roberto Navigli and Simone~Paolo Ponzetto. 2012.
\newblock \href{https://doi.org/10.1016/j.artint.2012.07.001}{{B}abel{N}et:
  {T}he automatic construction, evaluation and application of a wide-coverage
  multilingual semantic network}.
\newblock {\em Artificial Intelligence\/} 193:217--250.
\newblock
  \href{https://doi.org/10.1016/j.artint.2012.07.001}{https://doi.org/10.1016/j.artint.2012.07.001}.

\bibitem[{Nguyen et~al.(2016)Nguyen, Schulte~im Walde, and Vu}]{Nguyen:2016acl}
Kim~Anh Nguyen, Sabine Schulte~im Walde, and Ngoc~Thang Vu. 2016.
\newblock \href{http://anthology.aclweb.org/P16-2074}{Integrating
  distributional lexical contrast into word embeddings for antonym-synonym
  distinction}.
\newblock In {\em Proceedings of ACL\/}. pages 454--459.
\newblock
  \href{http://anthology.aclweb.org/P16-2074}{http://anthology.aclweb.org/P16-2074}.

\bibitem[{Osborne et~al.(2016)Osborne, Narayan, and Cohen}]{Osborne:16}
Dominique Osborne, Shashi Narayan, and Shay Cohen. 2016.
\newblock \href{https://arxiv.org/abs/1509.01007}{Encoding prior knowledge with
  eigenword embeddings}.
\newblock {\em Transactions of the ACL\/} 4:417--430.
\newblock
  \href{https://arxiv.org/abs/1509.01007}{https://arxiv.org/abs/1509.01007}.

\bibitem[{Pavlick et~al.(2015)Pavlick, Rastogi, Ganitkevitch, Durme, and
  Callison{-}Burch}]{Pavlick:2015acl}
Ellie Pavlick, Pushpendre Rastogi, Juri Ganitkevitch, Benjamin~Van Durme, and
  Chris Callison{-}Burch. 2015.
\newblock \href{http://www.aclweb.org/anthology/P15-2070}{{PPDB} 2.0: {B}etter
  paraphrase ranking, fine-grained entailment relations, word embeddings, and
  style classification}.
\newblock In {\em Proceedings of ACL\/}. pages 425--430.
\newblock
  \href{http://www.aclweb.org/anthology/P15-2070}{http://www.aclweb.org/anthology/P15-2070}.

\bibitem[{Pennington et~al.(2014)Pennington, Socher, and
  Manning}]{Pennington:2014emnlp}
Jeffrey Pennington, Richard Socher, and Christopher Manning. 2014.
\newblock \href{http://www.aclweb.org/anthology/D14-1162}{{Glove: G}lobal
  vectors for word representation}.
\newblock In {\em Proceedings of EMNLP\/}. pages 1532--1543.
\newblock
  \href{http://www.aclweb.org/anthology/D14-1162}{http://www.aclweb.org/anthology/D14-1162}.

\bibitem[{Perez and Liu(2017)}]{Perez:16b}
Julien Perez and Fei Liu. 2017.
\newblock \href{http://www.aclweb.org/anthology/E17-1029}{{Dialog state
  tracking, a machine reading approach using Memory Network}}.
\newblock In {\em Proceedings of EACL\/}. pages 305--314.
\newblock
  \href{http://www.aclweb.org/anthology/E17-1029}{http://www.aclweb.org/anthology/E17-1029}.

\bibitem[{Qiu et~al.(2014)Qiu, Cui, Bian, Gao, and Liu}]{Qiu:2014coling}
Siyu Qiu, Qing Cui, Jiang Bian, Bin Gao, and Tie-Yan Liu. 2014.
\newblock \href{http://www.aclweb.org/anthology/C14-1015}{Co-learning of word
  representations and morpheme representations}.
\newblock In {\em Proceedings of COLING\/}. pages 141--150.
\newblock
  \href{http://www.aclweb.org/anthology/C14-1015}{http://www.aclweb.org/anthology/C14-1015}.

\bibitem[{Schone and Jurafsky(2001)}]{Schone:2001naacl}
Patrick Schone and Daniel Jurafsky. 2001.
\newblock \href{http://aclweb.org/anthology/N/N01/N01-1024}{Knowledge-free
  induction of inflectional morphologies}.
\newblock In {\em Proceedings of NAACL\/}.
\newblock
  \href{http://aclweb.org/anthology/N/N01/N01-1024}{http://aclweb.org/anthology/N/N01/N01-1024}.

\bibitem[{Schwartz et~al.(2015)Schwartz, Reichart, and
  Rappoport}]{Schwartz:2015conll}
Roy Schwartz, Roi Reichart, and Ari Rappoport. 2015.
\newblock \href{http://www.aclweb.org/anthology/K15-1026}{Symmetric pattern
  based word embeddings for improved word similarity prediction}.
\newblock In {\em Proceedings of CoNLL\/}. pages 258--267.
\newblock
  \href{http://www.aclweb.org/anthology/K15-1026}{http://www.aclweb.org/anthology/K15-1026}.

\bibitem[{Schwartz et~al.(2016)Schwartz, Reichart, and
  Rappoport}]{Schwartz:2016naacl}
Roy Schwartz, Roi Reichart, and Ari Rappoport. 2016.
\newblock \href{http://www.aclweb.org/anthology/N16-1060}{Symmetric patterns
  and coordinations: {F}ast and enhanced representations of verbs and
  adjectives}.
\newblock In {\em Proceedings of NAACL-HLT\/}. pages 499--505.
\newblock
  \href{http://www.aclweb.org/anthology/N16-1060}{http://www.aclweb.org/anthology/N16-1060}.

\bibitem[{Soricut and Och(2015)}]{Soricut:2015naacl}
Radu Soricut and Franz Och. 2015.
\newblock \href{http://www.aclweb.org/anthology/N15-1186}{Unsupervised
  morphology induction using word embeddings}.
\newblock In {\em Proceedings of NAACL-HLT\/}. pages 1627--1637.
\newblock
  \href{http://www.aclweb.org/anthology/N15-1186}{http://www.aclweb.org/anthology/N15-1186}.

\bibitem[{Su et~al.()Su, Ga\v{s}i\'c, Mrk\v{s}i\'c, Rojas-Barahona, Ultes,
  Vandyke, Wen, and Young}]{su:2016:nnpolicy}
Pei-Hao Su, Milica Ga\v{s}i\'c, Nikola Mrk\v{s}i\'c, Lina Rojas-Barahona,
  Stefan Ultes, David Vandyke, Tsung-Hsien Wen, and Steve Young. ????
\newblock Continuously learning neural dialogue management.

\bibitem[{Su et~al.(2016)Su, Ga\v{s}i\'c, Mrk\v{s}i\'c, Rojas-Barahona, Ultes,
  Vandyke, Wen, and Young}]{Su:16}
Pei-Hao Su, Milica Ga\v{s}i\'c, Nikola Mrk\v{s}i\'c, Lina Rojas-Barahona,
  Stefan Ultes, David Vandyke, Tsung-Hsien Wen, and Steve Young. 2016.
\newblock \href{http://www.aclweb.org/anthology/P16-1230}{On-line active reward
  learning for policy optimisation in spoken dialogue systems}.
\newblock In {\em Proceedings of ACL\/}. pages 2431--2441.
\newblock
  \href{http://www.aclweb.org/anthology/P16-1230}{http://www.aclweb.org/anthology/P16-1230}.

\bibitem[{Sylak-Glassman et~al.(2015)Sylak-Glassman, Kirov, Yarowsky, and
  Que}]{Sylak:2015acl}
John Sylak-Glassman, Christo Kirov, David Yarowsky, and Roger Que. 2015.
\newblock \href{http://www.aclweb.org/anthology/P15-2111}{A
  language-independent feature schema for inflectional morphology}.
\newblock In {\em Proceedings of ACL\/}. pages 674--680.
\newblock
  \href{http://www.aclweb.org/anthology/P15-2111}{http://www.aclweb.org/anthology/P15-2111}.

\bibitem[{Tsarfaty et~al.(2010)Tsarfaty, Seddah, Goldberg, Kuebler, Versley,
  Candito, Foster, Rehbein, and Tounsi}]{Tsarfaty:10}
Reut Tsarfaty, Djam\'{e} Seddah, Yoav Goldberg, Sandra Kuebler, Yannick
  Versley, Marie Candito, Jennifer Foster, Ines Rehbein, and Lamia Tounsi.
  2010.
\newblock \href{http://www.aclweb.org/anthology/W10-1401}{Statistical parsing
  of morphologically rich languages {(SPMRL)} {W}hat, how and whither}.
\newblock In {\em Proceedings of the NAACL Workshop on Statistical Parsing of
  Morphologically-Rich Languages\/}. pages 1--12.
\newblock
  \href{http://www.aclweb.org/anthology/W10-1401}{http://www.aclweb.org/anthology/W10-1401}.

\bibitem[{Turian et~al.(2010)Turian, Ratinov, and Bengio}]{Turian:2010acl}
Joseph~P. Turian, Lev{-}Arie Ratinov, and Yoshua Bengio. 2010.
\newblock \href{http://www.aclweb.org/anthology/P10-1040}{Word representations:
  {A} simple and general method for semi-supervised learning}.
\newblock In {\em Proceedings of ACL\/}. pages 384--394.
\newblock
  \href{http://www.aclweb.org/anthology/P10-1040}{http://www.aclweb.org/anthology/P10-1040}.

\bibitem[{Turney and Pantel(2010)}]{Turney:2010jair}
Peter~D. Turney and Patrick Pantel. 2010.
\newblock \href{https://doi.org/10.1613/jair.2934}{From frequency to meaning:
  vector space models of semantics}.
\newblock {\em Journal of Artifical Intelligence Research\/} 37(1):141--188.
\newblock
  \href{https://doi.org/10.1613/jair.2934}{https://doi.org/10.1613/jair.2934}.

\bibitem[{Verwimp et~al.(2017)Verwimp, Pelemans, Van~hamme, and
  Wambacq}]{Verwimp:2017eacl}
Lyan Verwimp, Joris Pelemans, Hugo Van~hamme, and Patrick Wambacq. 2017.
\newblock \href{http://www.aclweb.org/anthology/E17-1040}{{Character-word LSTM
  language models}}.
\newblock In {\em Proceedings of EACL\/}. pages 417--427.
\newblock
  \href{http://www.aclweb.org/anthology/E17-1040}{http://www.aclweb.org/anthology/E17-1040}.

\bibitem[{Vodol\'{a}n et~al.(2017)Vodol\'{a}n, Kadlec, and
  Kleindienst}]{Vodolan:2017}
Miroslav Vodol\'{a}n, Rudolf Kadlec, and Jan Kleindienst. 2017.
\newblock \href{http://www.aclweb.org/anthology/E17-2033}{Hybrid dialog state
  tracker with {ASR} features}.
\newblock In {\em Proceedings of EACL\/}. pages 205--210.
\newblock
  \href{http://www.aclweb.org/anthology/E17-2033}{http://www.aclweb.org/anthology/E17-2033}.

\bibitem[{Vuli\'{c} and Korhonen(2016{\natexlab{a}})}]{Vulic:2016acluniversal}
Ivan Vuli\'{c} and Anna Korhonen. 2016{\natexlab{a}}.
\newblock \href{http://anthology.aclweb.org/P16-2084}{Is "universal syntax"
  universally useful for learning distributed word representations?}
\newblock In {\em Proceedings of ACL\/}. pages 518--524.
\newblock
  \href{http://anthology.aclweb.org/P16-2084}{http://anthology.aclweb.org/P16-2084}.

\bibitem[{Vuli\'{c} and Korhonen(2016{\natexlab{b}})}]{Vulic:2016acl}
Ivan Vuli\'{c} and Anna Korhonen. 2016{\natexlab{b}}.
\newblock \href{http://www.aclweb.org/anthology/P16-1024}{On the role of seed
  lexicons in learning bilingual word embeddings}.
\newblock In {\em Proceedings of ACL\/}. pages 247--257.
\newblock
  \href{http://www.aclweb.org/anthology/P16-1024}{http://www.aclweb.org/anthology/P16-1024}.

\bibitem[{Wang et~al.(2014)Wang, Zhang, Feng, and Chen}]{Wang:2014}
Zhen Wang, Jianwen Zhang, Jianlin Feng, and Zheng Chen. 2014.
\newblock Knowledge graph embedding by translating on hyperplanes.
\newblock In {\em Proceedings of AAAI\/}. pages 1112--1119.

\bibitem[{Wen et~al.(2017)Wen, Vandyke, Mrk{\v{s}}i\'c, Ga{\v{s}}i\'c,
  M.~Rojas-Barahona, Su, Ultes, and Young}]{Wen:17}
Tsung-Hsien Wen, David Vandyke, Nikola Mrk{\v{s}}i\'c, Milica Ga{\v{s}}i\'c,
  Lina M.~Rojas-Barahona, Pei-Hao Su, Stefan Ultes, and Steve Young. 2017.
\newblock \href{http://www.aclweb.org/anthology/E17-1042}{A network-based
  end-to-end trainable task-oriented dialogue system}.
\newblock In {\em Proceedings of EACL\/}.
\newblock
  \href{http://www.aclweb.org/anthology/E17-1042}{http://www.aclweb.org/anthology/E17-1042}.

\bibitem[{Wieting et~al.(2015)Wieting, Bansal, Gimpel, and
  Livescu}]{Wieting:2015tacl}
John Wieting, Mohit Bansal, Kevin Gimpel, and Karen Livescu. 2015.
\newblock From paraphrase database to compositional paraphrase model and back.
\newblock {\em {Transactions of the ACL}\/} 3:345--358.

\bibitem[{Wieting et~al.(2016)Wieting, Bansal, Gimpel, and
  Livescu}]{Wieting:2016emnlp}
John Wieting, Mohit Bansal, Kevin Gimpel, and Karen Livescu. 2016.
\newblock \href{https://aclweb.org/anthology/D16-1157}{{Charagram: E}mbedding
  words and sentences via character n-grams}.
\newblock In {\em Proceedings of EMNLP\/}. pages 1504--1515.
\newblock
  \href{https://aclweb.org/anthology/D16-1157}{https://aclweb.org/anthology/D16-1157}.

\bibitem[{Williams et~al.(2016)Williams, Raux, and Henderson}]{Williams:16}
Jason~D. Williams, Antoine Raux, and Matthew Henderson. 2016.
\newblock
  \href{http://dad.uni-bielefeld.de/index.php/dad/article/view/3685}{{The
  Dialog State Tracking Challenge series: A review}}.
\newblock {\em Dialogue \& Discourse\/} 7(3):4--33.
\newblock
  \href{http://dad.uni-bielefeld.de/index.php/dad/article/view/3685}{http://dad.uni-bielefeld.de/index.php/dad/article/view/3685}.

\bibitem[{Xu et~al.(2014)Xu, Bai, Bian, Gao, Wang, Liu, and Liu}]{Xu:2014}
Chang Xu, Yalong Bai, Jiang Bian, Bin Gao, Gang Wang, Xiaoguang Liu, and
  Tie-Yan Liu. 2014.
\newblock \href{https://doi.org/10.1145/2661829.2662038}{{RC-NET: A} general
  framework for incorporating knowledge into word representations}.
\newblock In {\em Proceedings of CIKM\/}. pages 1219--1228.
\newblock
  \href{https://doi.org/10.1145/2661829.2662038}{https://doi.org/10.1145/2661829.2662038}.

\bibitem[{Young(2010)}]{young:10}
Steve Young. 2010.
\newblock {Cognitive User Interfaces}.
\newblock {\em IEEE Signal Processing Magazine\/} .

\bibitem[{Yu and Dredze(2014)}]{Yu:2014}
Mo~Yu and Mark Dredze. 2014.
\newblock \href{http://www.aclweb.org/anthology/P14-2089}{Improving lexical
  embeddings with semantic knowledge}.
\newblock In {\em Proceedings of ACL\/}. pages 545--550.
\newblock
  \href{http://www.aclweb.org/anthology/P14-2089}{http://www.aclweb.org/anthology/P14-2089}.

\bibitem[{Zeller et~al.(2013)Zeller, \v{S}najder, and
  Pad\'{o}}]{Zeller:2013acl}
Britta Zeller, Jan \v{S}najder, and Sebastian Pad\'{o}. 2013.
\newblock \href{http://www.aclweb.org/anthology/P13-1118}{{DErivBase: I}nducing
  and evaluating a derivational morphology resource for {G}erman}.
\newblock In {\em Proceedings of ACL\/}. pages 1201--1211.
\newblock
  \href{http://www.aclweb.org/anthology/P13-1118}{http://www.aclweb.org/anthology/P13-1118}.

\bibitem[{Zilka and Jurcicek(2015)}]{Zilka:15}
Lukas Zilka and Filip Jurcicek. 2015.
\newblock Incremental {LSTM}-based dialog state tracker.
\newblock In {\em Proceedings of ASRU\/}.

\bibitem[{Zou et~al.(2013)Zou, Socher, Cer, and Manning}]{Zou:2013emnlp}
Will~Y. Zou, Richard Socher, Daniel Cer, and Christopher~D. Manning. 2013.
\newblock \href{http://www.aclweb.org/anthology/D13-1141}{Bilingual word
  embeddings for phrase-based machine translation}.
\newblock In {\em Proceedings of EMNLP\/}. pages 1393--1398.
\newblock
  \href{http://www.aclweb.org/anthology/D13-1141}{http://www.aclweb.org/anthology/D13-1141}.

\end{thebibliography}
\bibliographystyle{acl_natbib}


\clearpage



\twocolumn[\begin{center} 
        \Large {\bf Morph-fitting: Fine-Tuning Word Vector Spaces \\ with Simple Language-Specific Rules} \\ \textit{Supplementary Material} \\
        ~ \\
\end{center}]

\section*{Morphological Rules}
\label{a:rules}
In this supplemental material, we provide a short comprehensive overview of simple language-specific morphological rules in English (\textsc{en}), German (\textsc{de}), Italian (\textsc{ut}), and Russian (\textsc{ru}). These rules were used to build the sets of synonymous \textsc{Attract} and antonymous \textsc{Repel} constraints for our \textit{morph-fitting} fine-tuning procedure. As discussed in the paper, the linguistic constraints extracted from the rules require only a comprehensive list of vocabulary words in each language. A native speaker of each language used in our experiments is able to easily come up with these sets of morphological rules (or at least with a reasonable subset of rules) without any linguistic training. What is more, the rules for German, Italian, and Russian were created by non-native and non-fluent speakers who have only a passive or limited knowledge of the three languages, exemplifying the simplicity and portability of the fine-tuning approach based on the shallow ``morphological supervision''. The simplicity is also confirmed by the short time used to compile the rules, ranging from \textit{a few minutes} for English to approximately \textit{two hours} for Russian.

Different languages differ in their ``morphological richness'' (e.g., declension, verb conjugation, plural forming, gender) which consequently leads to the varying number of rules in each language. However, all four languages in our study display morphological regularities described by simple morphological rules that are exploited to build sets of \textsc{Attract} and \textsc{Repel} linguistic constraints in each language from scratch.\footnote{Note that the rules for extracting \textsc{Attract} constraints were additionally used to generate the Morph-SimLex evaluation set, also provided as supplemental material.} 

Vocabularies $W$ in all four languages are labeled $W_{en}$, $W_{de}$, $W_{it}$, $W_{ru}$. We add the pairs $(w_1, w_2)$ and $(w_2, w_1)$ generated by the rules to the sets of constraints iff both $w_1, w_2 \in W$. After we generate all such constraints, since some constraints may have been generated by more than one rule, we remove all duplicates from the respective sets of \textsc{Attract} and \textsc{Repel} constraints.

Before we start, we will define two simple functions: (i) the function $w[:-N]$ strips the last $N$ characters from the word $w$, (ii) the function $w$.\texttt{ew(sub)} tests if the word $w$ ends with a sequence of characters \texttt{sub}. For instance, $create[:-1]$ returns $creat$, while $create$.\texttt{ew('s')} returns \texttt{False} and $create$.\texttt{ew('e')} returns \texttt{True}.

\subsection*{English Rules}
\label{ss:english} 
\paragraph{Inflectional Synonymy: \textsc{Attract}}
As discussed in the paper, we rely on only two simple inflectional morphological rules in English: 

\noindent - $w_2 = w_1$ + \texttt{'s'/'ed'/'ing'}. This rule yields constraints such as \textit{(speak, speaking)}, \textit{(turtle, turtles)}, or \textit{(clean, cleaned)}.

\noindent - If $w_1$\texttt{.ew('e')}, then $w_2 = w_1[:-1]$ + \texttt{'ed'/'ing'}. This rule yields constraints such as \textit{(create, creating)}, or \textit{(generate, generated)}.

\paragraph{Derivational Antonymy: \textsc{Repel}}
We assume the following set of standard ``antonymy'' prefixes in English: $AP_{en}=$ \textit{\{'dis', 'il', 'un', 'in', 'im', 'ir', 'mis', 'non', 'anti'\}}. We rely on the following derivational rules to extract \textsc{Repel} pairs:

\noindent - $w_2$ = \texttt{ap} + $w_1$, where \texttt{ap} $\in AP_{en}$. This rule yields constraints such as \textit{(mature, immature)}, \textit{(allow, disallow)} or \textit{(regularity, irregularity)}.

\noindent - If $w_1$\texttt{.ew('ful')}, then $w_2$ = $w_1[:-3]$ + \texttt{'less'}. This rule yields constraints such as \textit{(cheerful, cheerless)}.

As mentioned in the paper, for all four languages we further expand the set of \textsc{Repel} constraints by transitively combining antonymy pairs with inflectional \textsc{Attract} pairs. In simple words, \textit{the friend of my enemy is my enemy.} This means that, given an \textsc{Attract} pair \textit{(allow, allows)} and a \textsc{Repel} pair \textit{(allow, disallow)}, we extract another \textsc{Repel} pair \textit{(allows, disallow)}.

\subsection*{German Rules}
\label{ss:english} 
\paragraph{Inflectional Synonymy: \textsc{Attract}}
Being morphologically richer than English, the German language naturally requires more rules to describe its (inflectional) morphological richness and variation. First, we capture the regular \textit{declension} of nouns and adjectives by the following heuristic:

\noindent - Generate a set of words $W_{w_1}=\{w_1,w_2|w_2 = w_1 + \text{'e'/'em'/'en'/'er'/'es'}\}$; take the Cartesian product on $W_{w_1} \times W_{w_1}$ and then exclude $(w_i, w_i)$ pairs with identical words. This rule generates pairs such as \textit{(schottisch, schottische)}, \textit{(schottischem, schottischen)}. 

The second set of rules describes regular \textit{verb morphology}, i.e., verb conjugation in the present and past tense, and the formation of regular past participles. This set of rules may be expressed as:

\noindent - If $w_1$\texttt{.ew('en')}, then $w_1'=w_1[:-2]$. If $w_1'$\texttt{.ew('t')}, then generate a set of words $W_{w_1}=\{w_1,w_2|w_2 = w_1 + \text{'e'/'st'/'ete'/'etest'/'etet'/'eten'}, w_2 = \text{'ge'} + w_1' + \text{'et'}\}$, else (if not $w_1'$\texttt{.ew('t')}), generate a set of words $W_{w_1}=\{w_1,w_2|w_2 = w_1 + \text{'e'/'st'/'te'/'test'/'tet'/'ten'}, w_2 = \text{'ge'} + w_1' + \text{'t'}\}$. We then take the Cartesian product on $W_{w_1} \times W_{w_1}$. Again, all pairs with identical words were discarded. This rule yields pairs such as \textit{(machen, machten)}, \textit{(mache, gemacht)}, \textit{(kaufst, kauft)}, or \textit{(arbeite, arbeitete)} and \textit{(arbeiten, gearbeitet)}.

Another set of rules targets the regular formation of \textit{plural} nouns:

\noindent - If $w_1$\texttt{.ew('ei')} or $w_1$\texttt{.ew('heit')} or $w_1$\texttt{.ew('keit')} or $w_1$\texttt{.ew('schaft')} or $w_1$\texttt{.ew('ung')}, then $w_2$ = $w_1$ + \texttt{'en'}. This rule yields pairs such as \textit{(wahrheit, wahrheiten)} or \textit{(gemeinschaft, gemeinschaften)}.

\noindent - If $w_1$\texttt{.ew('in')}, then $w_2$ = $w_1$ + \texttt{'nen'}. This rule generates pairs such as \textit{(lehrerin, lehrerinnen)} or \textit{(lektorin, lektorinnen)}.

\noindent - If $w_1$\texttt{.ew('a'/'i'/'o'/'u'/'y')} then $w_2$ = $w_1$ + \texttt{'s'}. This rule yields pairs such as \textit{(auto, autos)}.

\noindent - If $w_1$\texttt{.ew('e')}, then $w_2$ = $w_1$ + \texttt{'n'}. This rule yields pairs such as \textit{(postkarte, postkarten)}.

\noindent - $w_2 = lumlaut(w_1)$ + \texttt{er}, where the function $lumlaut(w)$ replaces the last occurrence of the letter \texttt{'a'},\texttt{'o'} or \texttt{'u'} with \texttt{'ä'},\texttt{'ö'} or \texttt{'ü'}. This rule generates pairs such as \textit{(wörterbuch, wörterbücher)} or \textit{(stadt, städter)}.

\paragraph{Derivational Antonymy: \textsc{Repel}}
We assume the following set of standard ``antonymy'' prefixes in German: $AP_{de}=$ \textit{\{'un', 'nicht', 'anti', 'ir', 'in', 'miss'\}}. We rely on the following derivational rules to extract \textsc{Repel} pairs in German:

\noindent - $w_2$ = \texttt{ap} + $w_1$, where \texttt{ap} $\in AP_{de}$. This rule yields constraints such as \textit{(aktiv, inaktiv)}, \textit{(wandelbar, unwandelbar)} or \textit{(zyklone, antizyklone)}.

\noindent - If $w_1$\texttt{.ew('voll')}, then $w_2$ = $w_1[:-4]$ + \texttt{'los'}. This rule yields constraints such as \textit{(geschmackvoll, geschmacklos)}.

The set of \textsc{Repel} is then again transitively expanded yielding pairs such as \textit{(relevant, irrelevanter)} or \textit{(aktivem, inaktiv)}.

\subsection*{Italian Rules}
\label{ss:italian}
\paragraph{Inflectional Synonymy: \textsc{Attract}}
The first set of rules aims at capturing the regular plural forming in Italian (e.g., \textit{libro, libri}) and regular differences in gender (e.g., \textit{rapido, rapida}). We rely on the simple heuristic which can be expressed as follows:

\noindent - If $w_1$\texttt{.ew('a'/'e'/'o'/'i')}, then generate a set of words $W_{w_1}=\{w_2|w_2 = w_1[:-1] + \text{'a'/'e'/'o'/'i'}\}$, and take the Cartesian product on $W_{w_1} \times W_{w_1}$ discarding pairs with identical words. This rule yields pairs such as \textit{(nero, neri)} or \textit{(generazione, generazioni)}.

\noindent - If $w_1$\texttt{.ew('ga'/'ca')}, then $w_2 = w_1$ + \texttt{'he'}. This rule generates pairs such as \textit{(tartaruga, tartarughe)} or \textit{(bianca, bianche)}.

\noindent - If $w_1$\texttt{.ew('go')}, then $w_2 = w_1$ + \texttt{'hi'}. This rule generates pairs such as \textit{(albergo, alberghi)}.

The second set of rules targets regular verb conjugation in Italian and the formation of regular past participles. The following rules are used:

\noindent - If $w_1$\texttt{.ew('are')}, then generate a set of words $W_{w_1}=\{w_1,w_2|w_2 = w_1[:-3] + \text{'iamo'/'ate'/'ano'/'o'/'i'/'a'/'ato'/'ata'/'ati'/'ate'}\}$; take the Cartesian product on $W_{w_1} \times W_{w_1}$ discarding pairs with identical words. This rule results in pairs such as \textit{(aspettare, aspettiamo)}.

\noindent - If $w_1$\texttt{.ew('ere')}, then generate a set of words $W_{w_1}=\{w_1,w_2|w_2 = w_1[:-3] + \text{'iamo'/'ete'/'ono'/'o'/'i'/'e'/'uto'/'uta'/'uti'/'ute'}\}$; take the Cartesian product on $W_{w_1} \times W_{w_1}$ discarding pairs with identical words. This rule results in pairs such as \textit{(ricevere, ricevete)} or \textit{(riceve, ricevuto)}.

\noindent - If $w_1$\texttt{.ew('ire')}, then generate a set of words $W_{w_1}=\{w_1,w_2|w_2 = w_1[:-3] + \text{'iamo'/'ite'/'ono'/'o'/'i'/'e'/'ito'/'ita'/'iti'/'ite'}\}$; take the Cartesian product on $W_{w_1} \times W_{w_1}$ discarding pairs with identical words. This rule results in pairs such as \textit{(dormire, dormono)} or \textit{(dormi, dormita)}.

\paragraph{Derivational Antonymy: \textsc{Repel}}
We assume the following set of standard ``antonymy'' prefixes: $AP_{it}=$ \textit{\{'in', 'ir', 'im', 'anti'\}}. The following derivational rule is used to extract \textsc{Repel} pairs:

\noindent - $w_2$ = \texttt{ap} + $w_1$, where \texttt{ap} $\in AP_{it}$. This rule yields constraints such as \textit{(attivo, inattivo)} or \textit{(rispettosa, irrispettosa)}.

The set of \textsc{Repel} was then expanded as before, e.g., with additional pairs such as \textit{(rispettosa, irrispettosi)} generated.

\vspace{-0.3em}
\subsection*{Russian Rules}
\paragraph{Inflectional Synonymy: \textsc{Attract}}
The first set of rules in Russian targets the regular forming of plural in Russian. A few simple heuristics are used as follows:

\noindent - $w_2 = w_1$ + \texttt{'\foreignlanguage{russian}{и}'/'\foreignlanguage{russian}{ы}'}. This rule yields pairs such as \foreignlanguage{russian}{(aльбом, aльбомы)}, transliterated as: \textit{(al'bom, al'bomy)}.

\noindent - if $w_1$\texttt{.ew('a'/'\foreignlanguage{russian}{я}'/'\foreignlanguage{russian}{ь}')}, then $w_2 = w_1[:-1]$ + \texttt{'\foreignlanguage{russian}{и}'/'\foreignlanguage{russian}{ы}'}. This rule generates pairs such as \foreignlanguage{russian}{(песня, песни)}: \textit{(pesnja, pesni)}.

\noindent - if $w_1$\texttt{.ew('o')}, then $w_2 = w_1[:-1]$ + \texttt{'\foreignlanguage{russian}{a}'}. This rule generates pairs such as \foreignlanguage{russian}{(письмо, письма)}: \textit{(pis'mo, pis'ma)}.

\noindent - if $w_1$\texttt{.ew('e')}, then $w_2 = w_1[:-1]$ + \texttt{'\foreignlanguage{russian}{я}'}. This rule generates pairs such as \foreignlanguage{russian}{(платье, платья)}: \textit{(plat'e, plat'ja)}.

The next set of rules targets regular verb conjugation of Russian verbs as well as the regular formation of past participles. We again build a simple heuristic to extract \textsc{Attract} pairs:

\noindent - if $w_1$\texttt{.ew('\foreignlanguage{russian}{ти}'/'\foreignlanguage{russian}{ть}')}, then generate a set of words $W_{w_1}=\{w_1,w_2|w_2 = w_1[:-2] + \text{\foreignlanguage{russian}{'у'/'ю'/'ешь'/'ишь'/'ет'/'ит'/'ем'/'им'}}, w_2 = w_1[:-2] + \text{\foreignlanguage{russian}{'ете'/ите'/'ут'/'ют'/'ат'/'ят'}}, w_2 = w_1[:-2] + \text{\foreignlanguage{russian}{'нный'/'нная'}}\}$ and take the Cartesian product on $W_{w_1} \times W_{w_1}$ discarding pairs with identical words. This rule yields pairs such as \foreignlanguage{russian}{(варить, варите)} or \foreignlanguage{russian}{(заканчиваю, заканчивают)}, transliterated as: \textit{(varit', varite)}, \textit{(zakanchivaju, zakanchivajut)}.

Following that, we also utilise the regularities regarding declension processes in Russian, captured by the following rules:

\noindent - if $w_1$\texttt{.ew(\foreignlanguage{russian}{'a'})}, then generate a set of words $W_{w_1} = \{w_1,w_2|w_2 = w_1[:-1] + \text{\foreignlanguage{russian}{'e'/'y'/'ой'}}\}$ and take the Cartesian product on $W_{w_1} \times W_{w_1}$ discarding pairs with identical words. This rule yields pairs such as \foreignlanguage{russian}{(работа, работой)}: \textit{(rabota, rabotoj)}.

\noindent - if $w_1$\texttt{.ew(\foreignlanguage{russian}{'я'})}, then generate a set of words $W_{w_1} = \{w_1,w_2|w_2 = w_1[:-1] + \text{\foreignlanguage{russian}{'e'/'ю'/'ей'}}\}$ and take the Cartesian product on $W_{w_1} \times W_{w_1}$ discarding pairs with identical words. This rule yields pairs such as \foreignlanguage{russian}{(линия, линию)}: \textit{(linija, liniju)}.

\noindent - if $w_1$\texttt{.ew(\foreignlanguage{russian}{'ы'})}, then generate a set of words $W_{w_1} = \{w_1,w_2|w_2 = w_1[:-1] + \text{\foreignlanguage{russian}{'ам'/'ами'/'ах'}}\}$ and take the Cartesian product on $W_{w_1} \times W_{w_1}$ discarding pairs with identical words. This rule yields pairs such as \foreignlanguage{russian}{(работам, работами)}: \textit{(rabotam, rabotami)}.

\noindent - if $w_1$\texttt{.ew(\foreignlanguage{russian}{'и'})}, then generate a set of words $W_{w_1} = \{w_1,w_2|w_2 = w_1[:-1] + \text{\foreignlanguage{russian}{'ь'/'ям'/'ями'/'ях'}}\}$ and take the Cartesian product on $W_{w_1} \times W_{w_1}$ discarding pairs with identical words. This rule yields pairs such as \foreignlanguage{russian}{(работам, работами)}: \textit{(rabotam, rabotami)}.

Yet another set of rules targets regular adjective comparison and gender:

\noindent - if $w_1$\texttt{.ew(\foreignlanguage{russian}{'ый'/'ой'/'ий'})}, then generate a set of words $W_{w_1} = \{w_1,w_2|w_2 = w_1[:-2] + \text{\foreignlanguage{russian}{'ь'/'ее'/'ые'}}\}$. This rule yields pairs such as \foreignlanguage{russian}{(быстрый, быстрее)}: \textit{(bystryj, bystree)}.

\noindent - if $w_1$\texttt{.ew(\foreignlanguage{russian}{'ая'})}, then generate a set of words $W_{w_1} = \{w_1,w_2|w_2 = w_1[:-2] + \text{\foreignlanguage{russian}{'ее'/'ыe'/'ый'}}\}$. This rule yields pairs such as \foreignlanguage{russian}{(новая, новыe)}: \textit{(novaja, novye)}.

\noindent - if $w_1$\texttt{.ew(\foreignlanguage{russian}{'oe'})}, then generate a set of words $W_{w_1} = \{w_1,w_2|w_2 = w_1[:-2] + \text{\foreignlanguage{russian}{'ый'/'ыe'/'ая'}}\}$. This rule yields pairs such as \foreignlanguage{russian}{(новое, новый)}: \textit{(novoe, novyj)}.

\paragraph{Derivational Antonymy: \textsc{Repel}}
We assume the following set of standard ``antonymy'' prefixes in Russian: $AP_{ru}=$ \textit{\{\foreignlanguage{russian}{не, анти}'\}}, and simply use the following rule:

\noindent - $w_2$ = \texttt{ap} + $w_1$, where \texttt{ap} $\in AP_{ru}$. This rule yields constraints such as \foreignlanguage{russian}{(адекватный, неадекватный)} or \foreignlanguage{russian}{(вирусная, антивирусная)}, transliterated as: \textit{(adekvatnyj, neadekvatnyj)} and \textit{(virusnaja, antivirusnaja)}.

The further expansion of \textsc{Repel} constraints yields pairs such as \foreignlanguage{russian}{(адекватный, неадекватная)}: \textit{(adekvatnyj, neadekvatnaja)}.

\subsection*{Further Discussion}
We stress that the listed rules for all four languages are \textit{non-exhaustive} and do not cover all possible inflectional and derivational morphological phenomena. More linguistic constraints may be extracted by resorting to more sophisticated rules covering finer-grained morphological processes (e.g., covering irregular plural forming or irregular verb conjugation and past participle forming, or non-standard declensions). Further, the listed rules, written by non-native speakers without any linguistic training in a very short time span, do not necessarily rely on established linguistic theories in each language, but are rather simple heuristics aiming to capture morphological regularities.

\end{document}